\documentclass[lettersize,journal]{IEEEtran}
\usepackage{amsmath,amsfonts}
\usepackage{algorithm}
\usepackage{array}
\usepackage[caption=false,font=normalsize,labelfont=sf,textfont=sf]{subfig}
\usepackage{textcomp}
\usepackage{stfloats}
\usepackage{url}
\usepackage{verbatim}
\usepackage{graphicx}
\usepackage{cite}
\usepackage[dvipsnames]{xcolor}
\usepackage{booktabs}
\usepackage{multirow}
\usepackage{times}
\usepackage{microtype}
\usepackage{epsfig}
\usepackage{caption}
\usepackage{float}
\usepackage{placeins}
\usepackage{color, colortbl}
\usepackage{stfloats}
\usepackage{enumitem}
\usepackage{tabularx}
\usepackage{xstring}
\usepackage{multirow}
\usepackage{xspace}
\usepackage{url}
\usepackage{subcaption}
\usepackage{xcolor}
\usepackage[hang,flushmargin]{footmisc}
\usepackage{algpseudocode}
\definecolor{cvprblue}{rgb}{0.21,0.49,0.74}
\usepackage[pagebackref,breaklinks,colorlinks,citecolor=cvprblue]{hyperref}
\usepackage{listings}
\usepackage{bm}

\newcommand*{\ie}{\emph{i.e.}\@\xspace}

\newcommand*{\vs}{\emph{v.s.}\@\xspace}
\makeatletter

\newcommand{\Rmnum}[1]{\expandafter\@slowromancap\romannumeral #1@}
\makeatother

\begin{document}

\title{DiffusionAD: Norm-guided One-step Denoising Diffusion for Anomaly Detection}

\author{Hui Zhang, Zheng Wang, Dan Zeng, Zuxuan Wu, and Yu-Gang Jiang~\IEEEmembership{Fellow,~IEEE} 
\thanks{
This work was supported in part by the National Natural Science Foundation of China under Grants 62032006 and 62302453. (Corresponding author:
Zuxuan Wu and Yu-Gang Jiang.)

H. Zhang, Z. Wu, Y-G. Jiang are with the Institute of Trustworthy Embodied AI, Fudan University. \protect  E-mail: hui\_zhang23@m.fudan.edu.cn, \{zxwu, ygj\}@fudan.edu.cn}
\thanks{Z. Wang is with the School of Computer Science, Zhejiang University of Technology. \protect Email: zhengwang@zjut.edu.cn}
\thanks{D. Zeng is with the School of Communication \& Information Engineering, Shanghai University. \protect Email: dzeng@shu.edu.cn}
}

\markboth{Journal of \LaTeX\ Class Files,~Vol.~14, No.~8, August~2021}%
{Zhang \MakeLowercase{\textit{et al.}}: DiffusionAD: Norm-guided One-step Denoising Diffusion for Anomaly Detection}


\maketitle

\begin{abstract}
Anomaly detection has garnered extensive applications in real industrial manufacturing due to its remarkable effectiveness and efficiency. 
However, previous generative-based models have been limited by suboptimal reconstruction quality, hampering their overall performance.
We introduce DiffusionAD, a novel anomaly detection pipeline comprising a reconstruction sub-network and a segmentation sub-network.
A fundamental enhancement lies in our reformulation of the reconstruction process using a diffusion model into a noise-to-norm paradigm. 
Here, the anomalous region loses its distinctive features after being disturbed by Gaussian noise and is subsequently reconstructed into an anomaly-free one. 
Afterward, the segmentation sub-network predicts pixel-level anomaly scores based on the similarities and discrepancies between the input image and its anomaly-free reconstruction.
Additionally, given the substantial decrease in inference speed due to the iterative denoising nature of diffusion models, we revisit the denoising process and introduce a rapid one-step denoising paradigm. This paradigm achieves hundreds of times acceleration while preserving comparable reconstruction quality.
Furthermore, considering the diversity in the manifestation of anomalies, we propose a norm-guided paradigm to integrate the benefits of multiple noise scales, enhancing the fidelity of reconstructions.
Comprehensive evaluations on four standard and challenging benchmarks reveal that DiffusionAD outperforms current state-of-the-art approaches and achieves comparable inference speed, demonstrating the effectiveness and broad applicability of the proposed pipeline. Code is released at {\small\url{https://github.com/HuiZhang0812/DiffusionAD}}.
\end{abstract}


\section{Introduction}
\label{sec:intro}

\IEEEPARstart{S}{imilar} to how human perception and visual systems work, anomaly detection involves identifying and locating anomalies with little to no prior knowledge about them. Over the past decades, anomaly detection has been a mission-critical task and a spotlight in the computer vision community due to its wide range of applications~\cite{zou2022spd,roth2022patchcore,zavrtanik2021draem,liu2023simplenet}.

\begin{figure}[t]
  \centering
  \includegraphics[width=1.0\linewidth]{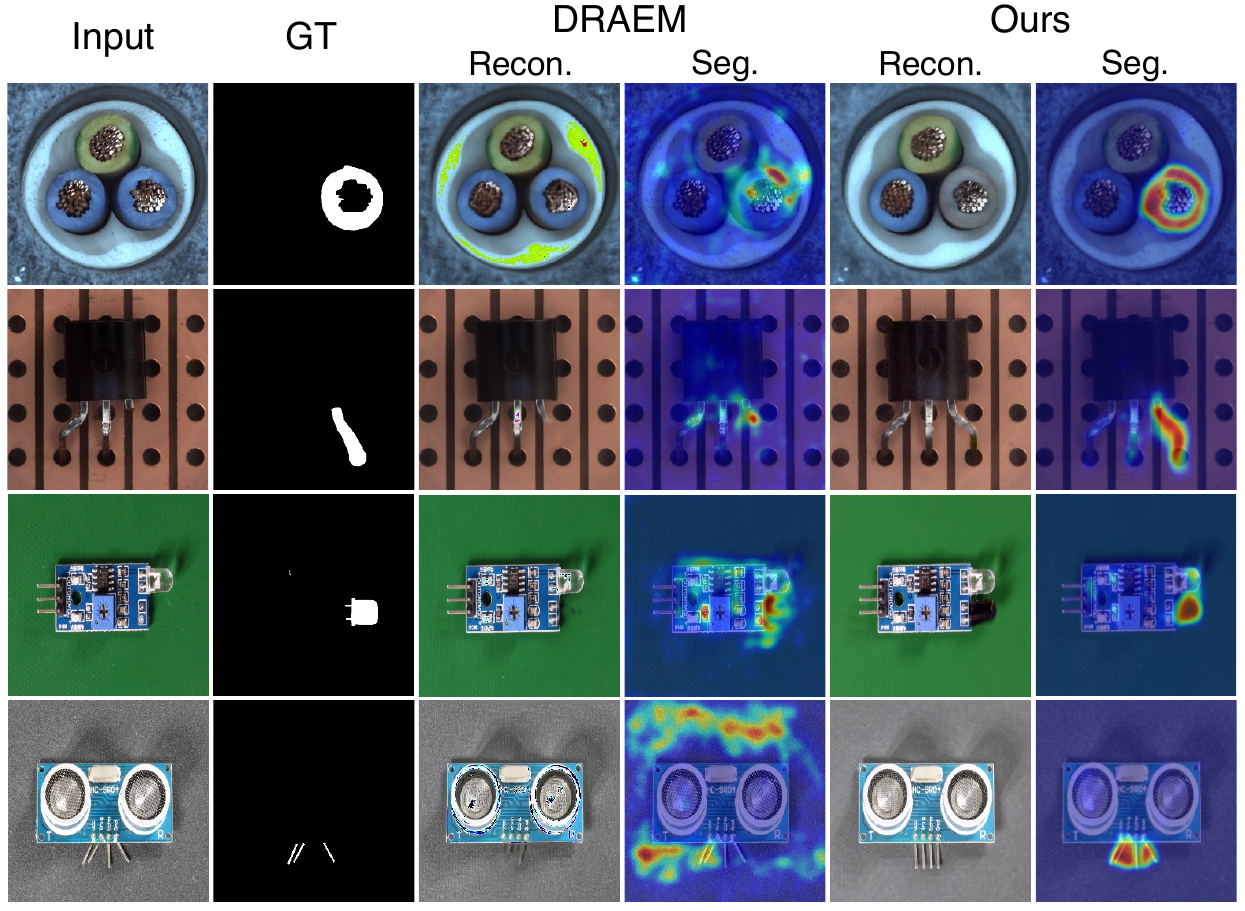}
  \caption{Anomaly detection and localization examples on MVTec~\cite{bergmann2019mvtec} and VisA~\cite{zou2022spd}. Compared to the previous autoencoder-based approach DRAEM~\cite{zavrtanik2021draem}, our proposed DiffusionAD exhibits superior reconstruction quality consisting of anomaly-free recovery of anomalous regions and fine-grained reconstruction of normal regions. Moreover, DiffusionAD locates the various anomaly regions more accurately.}
  \label{fig:main idea}
\end{figure}

Given its importance, a great number of work has been devoted to anomaly detection~(AD). Due to the limited number of anomaly samples and the labor-intensive labeling process, detailed anomaly samples are not available for training. As a result, most recent studies on anomaly detection have been performed without prior information about the anomaly, \ie, unsupervised paradigm~\cite{cohen2020spade,roth2022patchcore,deng2022rd4ad}. 
These methods include, but are not limited to, feature embeddings and generative models.
Feature embedding-based methods ~\cite{roth2022patchcore,deng2022rd4ad,cohen2020spade,liu2023simplenet,tien2023rd++}  often suffer from degraded performance when the distribution of industrial images differs significantly from the one used for feature extraction, as they rely on pre-trained feature extractors on extra datasets such as ImageNet.
Generative model-based methods ~\cite{akcay2018ganomaly,dehaene2020FAVAE,zavrtanik2021draem,ristea2022sspcab,liu2023dmad} require no extra data and are widely applicable in various scenarios. These approaches generally use autoencoder-based networks (AEs), based on the assumption that after the encoder has compressed the input image into a low-dimensional representation, the decoder will reconstruct the anomalous region as normal~\cite{baur2019medicalae,zavrtanik2021draem,gong2019memgan}.
However, as shown in Figure \ref{fig:main idea}, the AE-based paradigm has limitations: 
\Rmnum{1}) it may result in an \textbf{invariant reconstruction of abnormal regions} as the low-dimensional representation compressed from the original image still contains anomalous information, leading to false negative detection. 
\Rmnum{2}) AEs may perform a \textbf{coarse reconstruction of normal regions} due to limited restoration capability and introduce many false positives, especially on datasets with complex structures or textures. 

\IEEEpubidadjcol
To address the aforementioned issues, we propose a novel generative model-based framework consisting of a reconstruction sub-network and a segmentation sub-network for anomaly detection named DiffusionAD. 
Firstly, we reframe the reconstruction process as a \emph{noise-to-norm} paradigm by introducing Gaussian noise to perturb the input image, followed by a denoising model to predict the added noise. 
We implement noise addition and denoising via the diffusion model~\cite{ho2020ddpm} due to its excellent density estimation capability and high sampling quality.
The proposed paradigm offers two advantages, as shown in the fifth column of Figure \ref{fig:main idea}: 
\Rmnum{1}) The anomalous regions are treated as noise after losing their distinguishable features, which enables a \textbf{anomaly-free reconstruction} of the anomalous regions instead of an invariant one. 
\Rmnum{2}) It aims to cover the whole distribution of normal appearance~\cite{dhariwal2021diffusionbeatgans,ho2020ddpm}, which enables \textbf{fine-grained reconstruction} instead of a coarse reconstruction. 
After that, the segmentation sub-network predicts the pixel-wise anomaly score by exploiting the inconsistencies and commonalities between the input image and its reconstruction.

Equipped with the \emph{noise-to-norm} paradigm, DiffusionAD reconstructs more satisfactory results and thus improves the performance of anomaly detection.
However, as a class of likelihood-based models, diffusion models~\cite{song2020ddim,ho2020ddpm} generally require a large number of denoising iterations to obtain optimal reconstructions from randomly sampled Gaussian noise. 
Some prior methods~\cite{zhang2023diffad,lu2023KL_divergence,wyatt2022anoddpm} have made initial efforts to apply diffusion models to anomaly detection scenarios, indicating specific potential.
However, these methods require 20 to 1000 forward computations of the diffusion model network during the inference phase, which costs significant computational resources and is far slower than the real-time requirements.
To address this issue, we revisit the entire denoising process and introduce a \emph{one-step denoising} paradigm for anomaly detection that employs a diffusion model to predict the noise once and then directly predict the reconstruction result. 
This paradigm achieves dozens or even hundreds of times faster inference speed than the iterative denoising paradigm while maintaining comparable recovery quality.

\begin{figure}[t]
  \centering
  \includegraphics[width=1.0\linewidth]{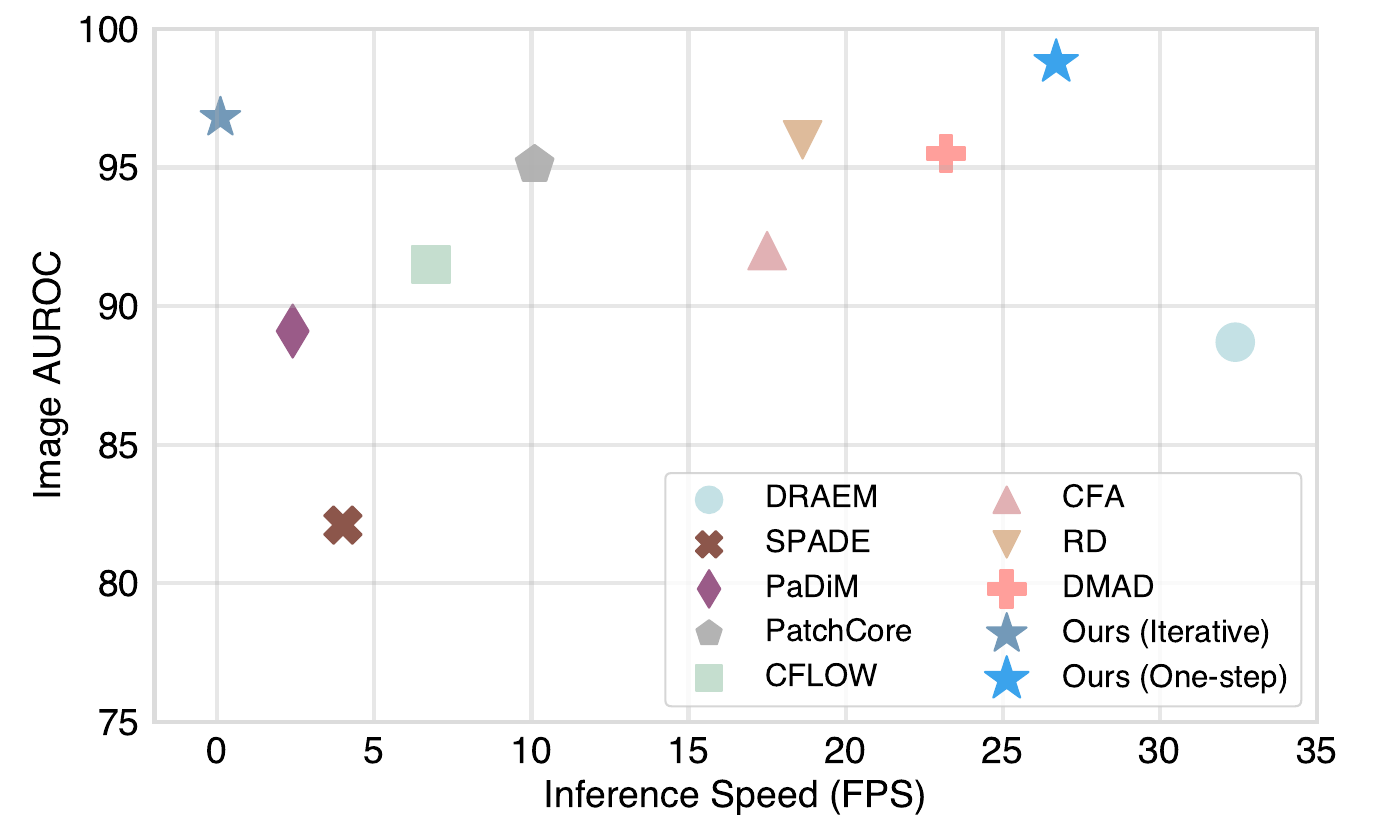}
  \caption{\textbf{Comparison of different algorithms on image AUROC and inference speed.} The Y-axis indicates the anomaly detection capability. The X-axis refers to the inference speed. These results are verified on the VisA~\cite{zou2022spd} dataset.}
  \label{fig:p-f-curves}
\end{figure}

Nonetheless, anomaly detection consistently poses a non-trivial challenge, mainly due to the inherent diversity in the representation of anomalies. These variants encompass subtle anomalies as well as more conspicuous ones, which are characterized by larger anomaly regions or semantic alterations.
We observe that different types of anomalies require different noise scales for efficient recovery, especially within the one-step denoising paradigm.
Specifically, when anomalies are relatively minor, a one-step prediction stemming from a smaller noise scale proves more advantageous, yielding superior pixel-level restoration quality.
For more pronounced anomalies, strategically applying a larger noise scale to perturb them and then reconstructing them via one-step denoising yields a higher quality of recovery at the semantic level. Therefore, we further introduce a \emph{norm-guided} paradigm that leverages direct predictions from larger noise scales to guide the reconstruction of smaller noise scales, leading to superior reconstruction outcomes.
Ultimately, DiffusionAD achieves SOTA anomaly detection performance and fast inference speed compared to other diffusion-free paradigms, as demonstrated in Figure \ref{fig:p-f-curves}.
This truly fulfills the effectiveness and efficiency requirements of real-world application scenarios.

The main contributions of this paper are summarized in the following:
\begin{itemize}

\item We propose DiffusionAD, a novel pipeline that reconstructs the input image to an anomaly-free one in real-time via the \emph{one-step noise-to-norm} paradigm and then predicts pixel-wise anomaly scores by exploiting inconsistencies and commonalities between them. 

\item We propose the \emph{norm-guided} paradigm to integrate the advantages of various noise scales in repairing different types of anomalies, aiming to achieve optimal anomaly-free reconstruction.

\item We conduct comprehensive experiments on four datasets to demonstrate that DiffusionAD significantly outperforms the previous SOTA by a large margin in terms of anomaly detection and localization.

\end{itemize}

\section{Related Work}
\label{sec:related}

\subsection{Anomaly Detection.\quad}Modern methods for anomaly detection encompass two main paradigms: feature embedding-based approaches~\cite{roth2022patchcore,deng2022rd4ad,cohen2020spade,defard2021padim,lee2022cfa,li2021cutpaste,zhang2023prn,liu2023simplenet,tien2023rd++} and generative model-based approaches~\cite{gudovskiy2022cflow,yu2021fastflow,rudolph2022csflow,zavrtanik2021draem,ristea2022sspcab}.

Methods based on feature embeddings typically extract features of normal samples through a model pre-trained on ImageNet and then perform anomaly estimation. Built on top of extracted features, knowledge distillation~\cite{bergmann2020us,salehi2021kdad,deng2022rd4ad,tien2023rd++} estimate anomalies by comparing the differences in anomaly region features between teacher and student networks. There are 
 also extensive studies ~\cite{cohen2020spade,defard2021padim,liu2023simplenet,lee2022cfa} estimating anomalies by measuring the distance between an anomalous sample and the feature space of normal samples.

On the other hand, generative model-based methods do not require additional data. The core idea of generative model-based approaches is to implicitly or explicitly learn the feature distribution of the anomaly-free training data. Generative models based on VAE~\cite{dehaene2019gvae,liu2020avae} introduce a multidimensional normal distribution in the latent space for normal samples and then estimate the anomaly by the negative log-likelihood of the established distribution. GAN-based generative models~\cite{schlegl2017gan,akcay2018ganomaly,yu2020cyclegan,gong2019memgan,hou2021dividegan} estimate anomalies through a discriminative network that compares the query image with randomly sampled samples from the latent space of the generative network. Besides, several works introduce proxy tasks based on the generative paradigm, such as image inpainting~\cite{zavrtanik2021draem,song2021anoseg} and attribute prediction~\cite{ristea2022sspcab,ye2020attribute,ulutas2020attribute}.
Normalizing Flow-based methods~\cite{rudolph2021differnet,gudovskiy2022cflow,rudolph2022csflow,yu2021fastflow} combine deep feature embeddings and generative models. These methods estimate accurate data likelihoods in the latent space by learning bijective transformations between normal sample distributions and specified densities.

\subsection{Diffusion Model.\quad}Diffusion models~\cite{ho2020ddpm,song2020ddim,song2020scoregm}, a class of generative models inspired by non-equilibrium thermodynamics~\cite{sohl2015nonequilibrium}, define a paradigm in which the forward process slowly adds random noise to the data, and the reverse constructs the desired data samples from the noise. Recently, a wide range of diffusion-based perception applications has emerged, such as image generation~\cite{dhariwal2021diffusionbeatgans,ho2020ddpm,song2020scoregm,rombach2022ldm}, image segmentation~\cite{kim2022diffusionseg,baranchukdiffusionseg,graikos2022diffusionseg,brempong2022denoisingseg}, object detection~\cite{chen2023diffusiondet}, etc. 
AnoDDPM~\cite{wyatt2022anoddpm} tentatively explores the application of diffusion models to reconstruct medical lesions in the brain.  
However, it measures unhealthy outliers by squared error, leading to a high false positive rate. 
In addition, the iterative denoising approach employed in AnoDDPM leads to a notably slow inference speed and substantial computational cost.
Some concurrent works~\cite{lu2023KL_divergence,zhang2023diffad} have attempted to utilize diffusion models to repair anomalous regions, demonstrating their potential feasibility. However, their inference speed still cannot meet the requirements of practical applications, even when using faster samplers.

\section{Preliminaries}
\label{sec:preliminaries}
\subsection{Denoising diffusion models.\quad}A family of generative models called denoising diffusion models~\cite{ho2020ddpm,song2020ddim,rombach2022ldm,sohl2015nonequilibrium} are inspired by equilibrium thermodynamics~\cite{song2020scoregm,song2019gradientsgm}. In particular, a diffusion probabilistic model specifies a forward diffusion phase in which the input data are progressively perturbed by adding Gaussian noise over several steps and then learns to reverse the diffusion process to recover the desired noise-free data from the noisy data. 
The forward noise process is defined as:
\begin{equation}
\begin{aligned}
    x_{t}=x_{0} \sqrt{\bar{\alpha}_{t}}+\epsilon_{t} \sqrt{1-\bar{\alpha}_{t}}, \quad \epsilon_{t} \sim \mathcal{N}(\mathbf{0}, \mathbf{I}),
\end{aligned}
\label{eq:xt}
\end{equation}
which transforms a data sample $x_{0}$ into a noisy sample $x_{t}$. Here, $t$ is randomly sampled from $\{0, 1, ... , T\}$, $\bar{\alpha}_{t}=\prod_{i=0}^{t} \alpha_{i}=\prod_{i=0}^{t}(1-\beta_{i})$ and $\beta_{i} \in(0,1)$ represents the noise variance schedule. This can be defined as a small linear schedule~\cite{sohl2015nonequilibrium} from $\beta_{1}=10^{-4}$ to $\beta_{T}=10^{-2}$. During training, a U-Net~\cite{ronneberger2015U-Net}-like architectures $\epsilon_{\theta}\left(x_{t}, t\right)$ is trained to predict $\epsilon$ by minimizing the training objective:
\begin{equation}
\begin{aligned}
\mathcal{L}=\mathbb{E}_{t \sim[1-T], x_{0} \sim q\left(x_{0}\right), \epsilon \sim \mathcal{N}(0, \mathbf{I})}\left[\left\|\epsilon-\epsilon_{\theta}\left(x_{t}, t\right)\right\|^{2}\right].
\end{aligned}
\label{eq:diffusion loss}
\end{equation}
At inference stage, $x_{t-1}$ is reconstructed from noise $x_{t}$ with the model $\epsilon_{\theta}\left(x_{t}, t\right)$ according to:
\begin{equation}
\begin{aligned}
x_{t-1}=\frac{1}{\sqrt{\alpha_{t}}}\left(x_{t}-\frac{1-\alpha_{t}}{\sqrt{1-\bar{\alpha}_{t}}} \epsilon_{\theta}\left(x_{t}, t\right)\right)+\tilde{\beta}_{t} z.
\end{aligned}
\label{eq:xt-1}
\end{equation}
where $z \sim \mathcal{N}(\mathbf{0}, \mathbf{I})$ and $\tilde{\beta}_{t}=\frac{1-\bar{\alpha}_{t-1}}{1-\bar{\alpha}_{t}} \beta_{t}$. $x_{0}$ is reconstructed from $x_{t}$ in an iterative way, 
\ie, $ x_{t} \rightarrow x_{t-1} \rightarrow \ldots \rightarrow x_{0}$.

\subsection{Classifier guidance and image guidance.\quad}To improve the quality and diversity of generated samples, guided diffusion~\cite{dhariwal2021diffusionbeatgans} uses a pre-trained classifier $p_\phi(y|x_t)$ to guide the diffusion sampling process, where $y$ is the class label. The guiding process is defined as modifying the noise prediction by a guidance scale $w$:
\begin{equation}
\begin{aligned}
\epsilon_\theta(x_t,t,y)=\epsilon_\theta(x_t,t)-\sqrt{1-\bar{\alpha}_t}\mathrm{~}w\nabla_{x_t}\log p_\phi(y|x_t)
\end{aligned}
\label{eq:classifier guidance}
\end{equation}
Thus, the impact of class $y$ on the generated results can be controlled by adjusting the parameter $w$.

To achieve a more diverse form of control, SDG~\cite{liu2023morecontrol} utilizes the reference image $r$ to guide the sampling process and reformulates (\ref{eq:classifier guidance}) as follows:
\begin{equation} 
\begin{aligned}
\epsilon_\theta({x_t,t,r_t)=\epsilon_\theta(x_t,t)-\sqrt{1-\bar{\alpha}_t}\mathrm{~}w\nabla_{x_t}\operatorname{sim}(x_t,r_t)}
\end{aligned}
\label{eq:image guidance}
\end{equation}
where $r_t$ is obtained from (\ref{eq:xt}) by perturbing $r$ and $\operatorname{sim}( \cdot, \cdot )$ is a measure of the similarity or correlation between two images.
In this paper, we adopt squared Euclidean distance to measure the similarity between two images.

\section{Methodology}
\label{sec:methodology}

\begin{figure*}
  \centering
  \includegraphics[width=0.98\linewidth]{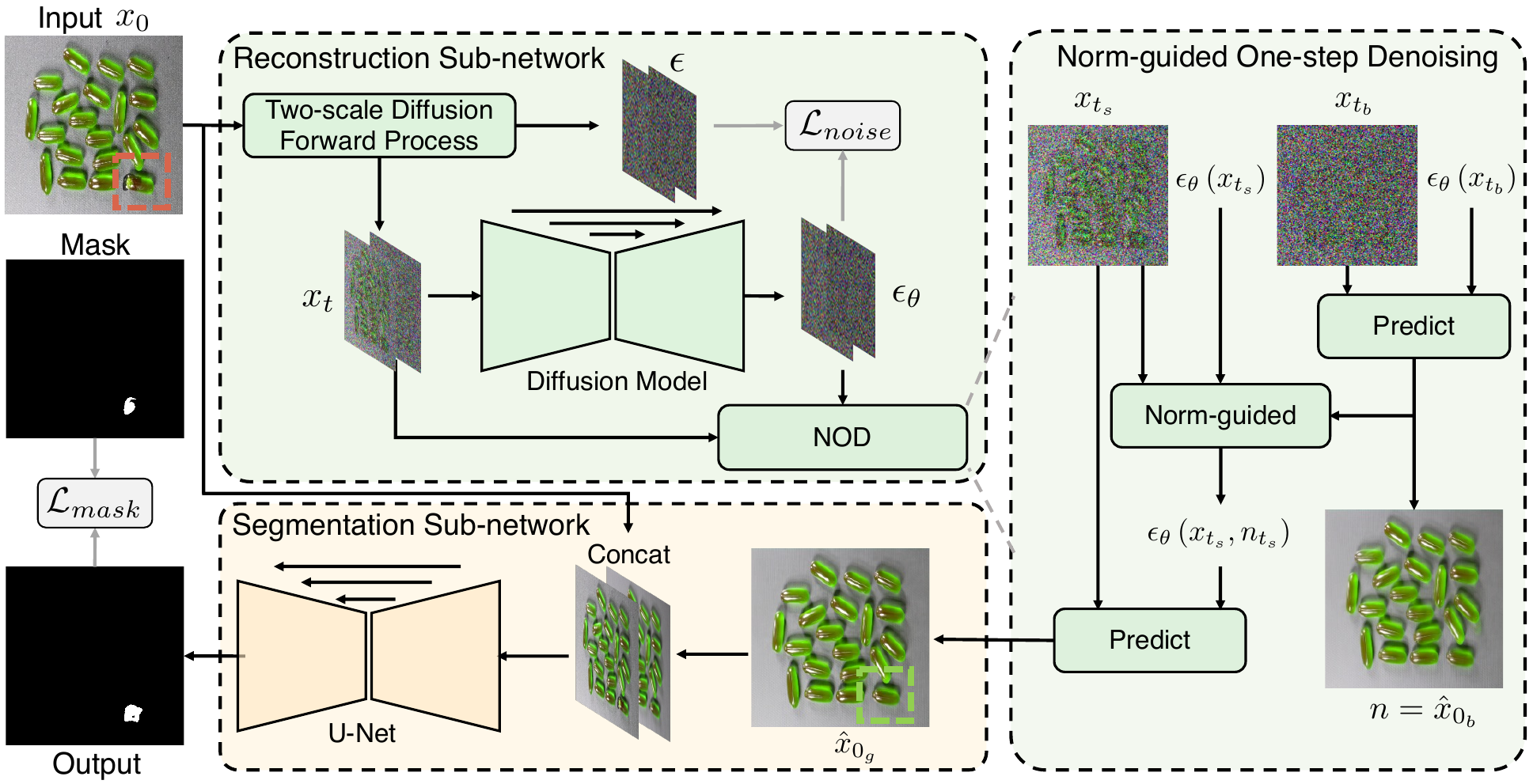}
  \caption{\textbf{An overview of the proposed pipeline DiffusionAD.} The reconstruction and segmentation sub-networks constitute the entire pipeline. 
  The input image is perturbed by two distinct noise scales. Following, the noise is predicted by an inference within the diffusion model. Finally, the norm-guided one-step denoising paradigm is employed to predict an anomaly-free reconstruction.
  The segmentation sub-network predicts pixel-wise anomaly scores by comparing commonalities and inconsistencies between the input image and its reconstruction.
  }
  \label{fig:architecture}
\end{figure*}

\subsection{Architecture}
\label{subsec:architecture}

Our proposed novel generative model-based framework consists of two components: a reconstruction sub-network and a segmentation sub-network, as shown in Figure \ref{fig:architecture}.
With the proposed anomaly synthesis strategies~(\ref{subsec:Anomaly Synthetic Strategy}), we define $x_0$ as the input that is either normal ($y=0$) or anomalous ($y=1$). 

\noindent\textbf{Reconstruction sub-network.\quad}We implement the reconstruction sub-network via a diffusion model, which reformulates the reconstruction process as a \emph{noise-to-norm} paradigm. 
First, we use the diffusion forward process proposed~\cite{ho2020ddpm} to corrupt the input image $x_0$ at a random time step $t$ to obtain $x_t$ via (\ref{eq:xt}). 
The input image $x_0$ gradually loses its discriminative features and approaches an isotropic Gaussian distribution as the time step increases. 
Then $\epsilon_{\theta}\left(x_{t}, t\right)$ is a function approximator intended to predict noise $\epsilon$ from ${x}_{t}$ and $t$, which is implemented with a U-Net~\cite{ronneberger2015U-Net,dhariwal2021diffusionbeatgans}-like architecture based on PixelCNN~\cite{salimanspixelcnn++}, ResNet~\cite{he2016resnet-18}, and Transformer~\cite{vaswani2017attention}. 
It is worth noting that after being perturbed by Gaussian noise, the anomalous pixels lose their distinctive features and tend to be treated by the model as injected noise. Consequently, the anomaly-free reconstruction $\hat{x}_{0}$ can be obtained from (\ref{eq:xt-1}) in an iterative manner.

\noindent\textbf{Segmentation sub-network.\quad}The segmentation sub-network employs a U-Net~\cite{ronneberger2015U-Net}-like architecture consisting of an encoder, a decoder, and skip connections. The input to the segmentation sub-network is a channel-wise concatenation of $x_{0}$ and $\hat{x}_{0}$. 
The segmentation sub-network learns to identify anomalies by exploiting the inconsistencies and commonalities between the input image $x_{0}$ and its anomaly-free approximation $\hat{x}_{0}$ to predict the pixel-wise anomaly score without post-processing. Remarkably, the learned inconsistency reduces false positives caused by slight pixel-wise differences between the normal region and its reconstruction and highlights significantly different regions.

\subsection{Norm-guided One-step Denoising}
Each iteration of the denoising process (\ref{eq:xt-1}) in the diffusion model corresponds to a round of network inference, requiring substantial computational resources and time. This presents a significant challenge for real-time inference. We revisit the entire denoising process, aiming to identify specific properties that enhance inference efficiency.

\begin{figure*}
  \centering
  \includegraphics[width=0.9\linewidth]{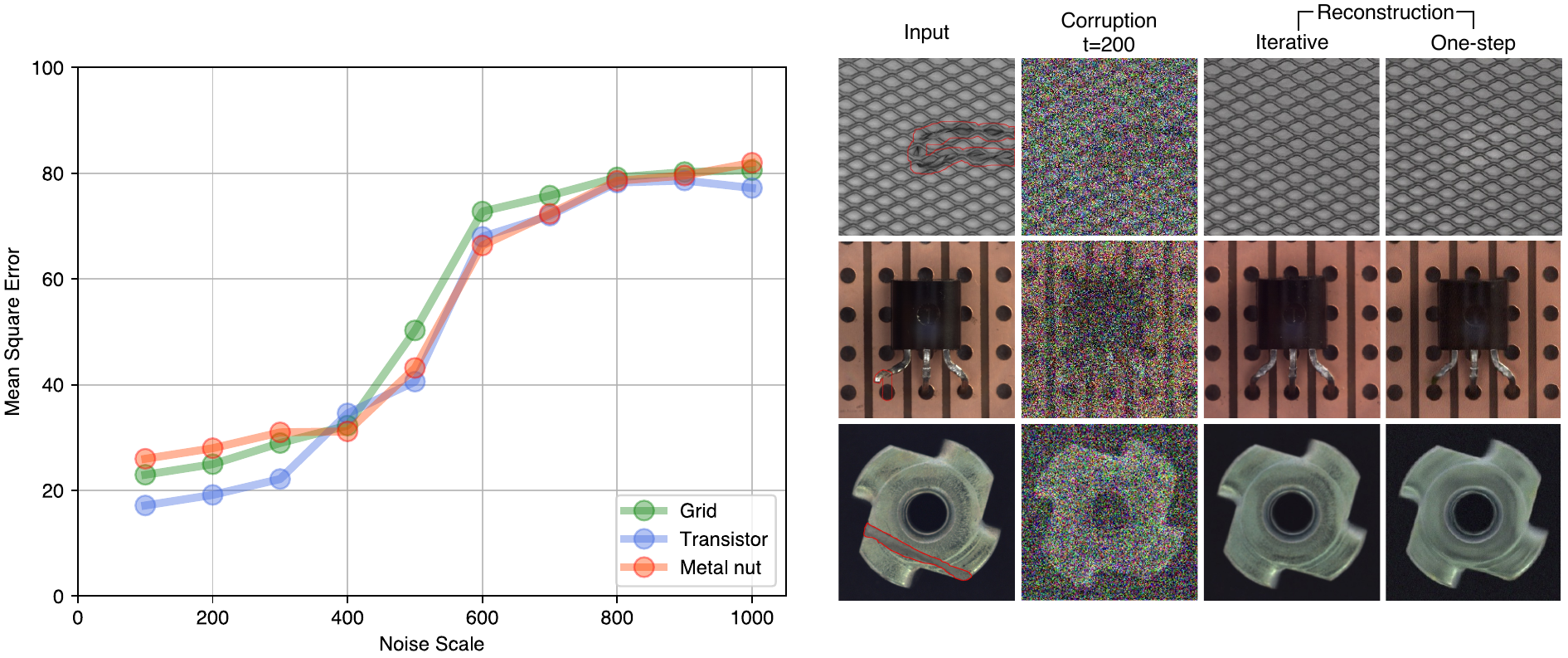}
  \caption{\textbf{Observation \Rmnum{1}.} When the noise scale is less than 500, the anomaly-free reconstruction results obtained through one-step denoising are comparable to those obtained through iterative denoising, as indicated by their low mean square error (left) and similar perceptual quality (right).}
  \label{fig:Observation1}
\end{figure*}

\noindent\textbf{Observation \Rmnum{1}.\quad}\emph{When the noise scale is small, the anomaly-free reconstruction via one-step denoising is comparable to the iterative one.}

At any time step $t$ in the denoising process, after the diffusion model predicts the noise $\epsilon_{\theta}\left(x_{t}, t\right)$ of $x_t$ by a single network inference, directly predict $\hat{x_0}$ is always valid~\cite{ho2020ddpm}. In this paper, we refer to the method of obtaining anomaly-free reconstructions $\hat{x}_{0}$ using only a single forward pass of the diffusion network as the one-step denoising paradigm.
In Figure \ref{fig:Observation1}, we provide empirical evidence related to observation \Rmnum{1}.
The experiments elucidate two main insights:
1) When using images with minor noise perturbations, the one-step denoising paradigm is comparable to the iterative denoising paradigm. As shown on the left side of figure~\ref{fig:Observation1}, the mean square errors (MSE) between the one-step and iterative reconstructions across three datasets remain consistently low when $t \leq 400$. The visualizations on the right side of figure~\ref{fig:Observation1} further demonstrate the similarity in perceptual quality between the one-step and iterative paradigms.
2) With more significant noise perturbations, such as $t \geq 500$, the results of one-step denoising generally exhibit significant distortion, reflected in higher MSE values.

\noindent\textbf{One-step Denoising.\quad}To this end, we employ one-step denoising as an alternative to the iterative denoising paradigm when the noise scale is less than 500.
More formally, given $\epsilon_{\theta}\left(x_{t}, t\right)$ and $x_t$, one-step reconstruction $\hat{x}_{0}$ can be obtained from the following equation:
\begin{equation}
\begin{aligned}
\hat{x}_{0}=\frac{1}{\sqrt{\bar{\alpha}_{t}}}\left(x_{t}-\sqrt{1-\bar{\alpha}_{t}} \epsilon_{\theta}\left(x_{t}, t\right)\right).
\end{aligned}
\label{eq:xhat}
\end{equation}
This paradigm is $t$ times faster than the iterative one, achieving significant savings in computational resources and inference time while maintaining comparable reconstruction quality.

\begin{figure}
  \centering
  \includegraphics[width=1.0\linewidth]{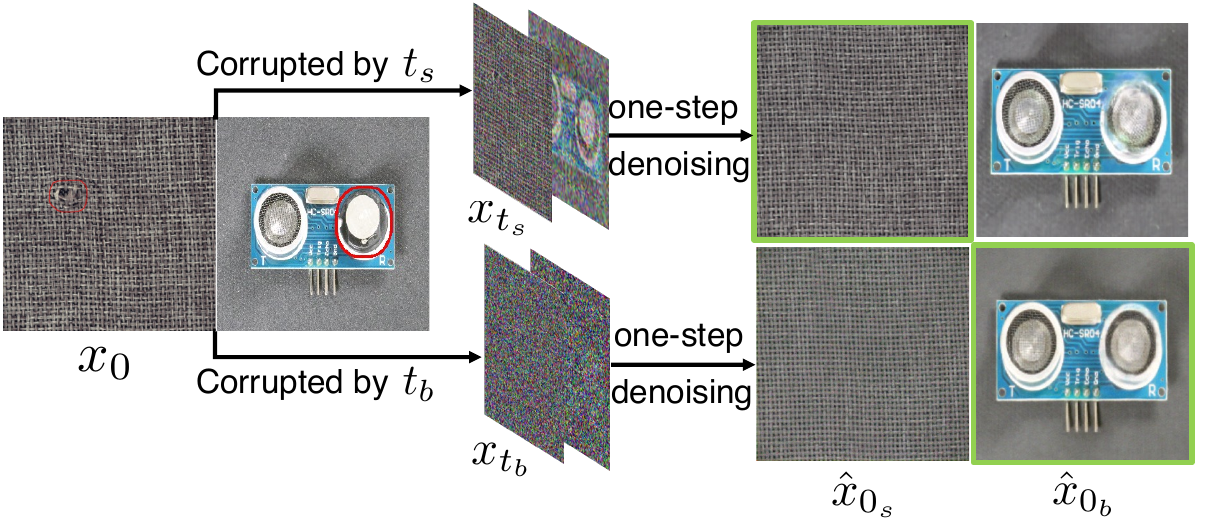}
  \caption{\textbf{Observation \Rmnum{2}.} The impact of two different noise scales ($t_s=200$ and $t_b=400$) on the anomaly-free reconstruction for different types of anomalies. Images with green borders represent superior reconstructions.}
  \label{fig:Observation2}
\end{figure}

\noindent\textbf{Observation \Rmnum{2}.\quad}\emph{Repairing different types of anomalies requires different scales of noise injection.}

In Figure \ref{fig:Observation2}, we present empirical experiments pertaining to this observation. 
To investigate the noise scales required for repairing different types of anomalies, we employ two distinct noise scales $t_s=200$ and $t_b=400$ to observe their impact on the anomaly-free reconstructions.
The experiments elucidate the following two insights:
1) When the anomaly region is relatively small, \ie small hole in the carpet, one-step denoising from $x_{t_s}$ is more suitable. The anomaly-free restoration $\hat{x}_{0_s}$ demonstrates higher perceptual quality, preserving fine-grained details. Conversely, $\hat{x}_{0_b}$ predicted directly from $x_{t_b}$ introduces some distortions.
2) In cases where the anomaly region is larger or exhibits semantic changes, \ie large white blockage on the circuit board, $\hat{x}_{0_s}$ predicted from $x_{t_s}$ exits residual anomaly regions. Therefore, the injection of noise at larger scales becomes necessary to effectively perturb the anomalous regions in order to achieve semantically high-quality anomaly-free restoration $\hat{x}_{0_b}$.
Thus, both of these noise scales are indispensable, as the appearance of anomalies varies significantly, \ie, abnormal regions can take on a variety of sizes, shapes, and numbers.  

\noindent\textbf{Norm-guided Denoising.\quad}To this end, in order to harness the advantages of both noise scale regimes, we propose a norm-guided denoising paradigm as shown in Figure \ref{fig:architecture}.
We divide the random range of $t\in \{0, 1, ... , T\}$ into two parts using $\tau$, where $S = \{0, 1, ... , \tau\}$ and $B= \{ \tau+1, \tau+2, ... , T\}$. 
For an input image $x_0$, we first perturb it using two randomly sampled time steps $t_s \in S $ and $t_b \in B $ to obtain $x_{t_s}$ and $x_{t_b}$ through (\ref{eq:xt}).
As the diffusion model defaults to being conditioned on $t$, we abbreviate $\epsilon_\theta(x_t,t)$ as $\epsilon_\theta(x_t)$.
Subsequently, we first employ the diffusion model to individually predict the noise of $x_{t_s}$ and $x_{t_b}$, corresponding to $\epsilon_\theta(x_{t_s})$ and $\epsilon_\theta(x_{t_b})$, respectively. We then directly predict $\hat{x}_{0_b}$ using $\epsilon_\theta(x_{t_b})$ by (\ref{eq:xhat}).  
Although $\hat{x}_{0_b}$ exhibits some distortions and appears low in mass, it consistently presents an anomaly-free appearance despite the diversity of anomalies in $x_0$. 
Hence, $\hat{x}_{0_b}$ is regarded as $\hat{x}_{0_b} \sim q(x_{y=0})$ and is rewritten as $n$.
After that, $n$ assumes the role of a conditional image to guide the prediction of $\hat{x}_{0_s}$ by (\ref{eq:image guidance}). 
We define $\operatorname{sim}(x_t, n_t) = -\frac{1}{2} \|n_{t_s} - x_{t_s}\|^2$ to measure the similarity between these two images.
$n_{t_s}$ is obtained by perturbing $n$ with the predicted noise $\epsilon_\theta(x_{t_s})$ instead of random noise in order to make the guidance more dependent on the difference between the image contents themselves.
Thus, (\ref{eq:image guidance}) is reformulated as follows:
\begin{equation} 
\begin{aligned}
\epsilon_\theta(x_{t_s},n_{t_s})=\epsilon_\theta(x_{t_s})-\sqrt{1-\bar{\alpha}_{t_s}}\mathrm{~}w\mathrm{~}(n_{t_s} - x_{t_s})
\end{aligned}
\label{eq:norm guidance}
\end{equation}
where $\epsilon_\theta(x_{t_s},n_{t_s})$ denotes the modified noise with norm-guidance. 
Finally, bring $x_{t_s}$ and $\epsilon_\theta(x_{t_s},n_{t_s})$ into (\ref{eq:xhat}) to directly predict the norm-guided anomaly-free reconstruction as follows:
\begin{equation}
\begin{aligned}
\hat{x}_{0_g}=\frac{1}{\sqrt{\bar{\alpha}_{t_s}}}\left(x_{t_s}-\sqrt{1-\bar{\alpha}_{t_s}} \epsilon_{\theta}\left(x_{t_s},n_{t_s} \right)\right).
\end{aligned}
\label{eq:guided xhat}
\end{equation}
We use the $\hat{x}_{0_b}$ to guide the denoising process of $\hat{x}_{0_s}$ in order to make the final reconstruction result $\hat{x}_{0_g}$ inherits both semantic correctness from $\hat{x}_{0_b}$ and detail fidelity from $\hat{x}_{0_s}$.
With this norm-guided denoising paradigm, DiffusionAD can handle different types of anomalies and predict superior reconstructions.

\subsection{Training \& Inference}
\label{subsec:training}

\noindent\textbf{Training stage.\quad}We jointly train the reconstruction and segmentation sub-networks.
The reconstruction sub-network is implemented as a diffusion model trained exclusively on normal (anomaly-free) samples (\ie, $y=0$), which explicitly excludes anomalies (\ie, $y=1$). It learns the entire distribution of normal patterns by minimizing the following loss function:

\begin{equation}
\begin{aligned}
\mathcal{L}_{noise}=\frac{(1-y)(\|\epsilon_{t_s}-\epsilon_{\theta}(x_{t_s})\|^{2} + \|\epsilon_{t_b}-\epsilon_{\theta}(x_{t_b})\|^{2})}{2}.
\end{aligned}
\label{eq:noise loss}
\end{equation}
The segmentation sub-network exploits the commonalities and differences between $x_{0}$ and $\hat{x}_{0_g}$ to predict pixel-wise anomaly scores as close as possible to the ground truth mask. The segmentation loss is defined as:
\begin{equation}
\begin{aligned}
\mathcal{L}_{mask}=\operatorname{Smooth}_{\mathcal{L} 1}\left(M, \hat{M}\right)+\gamma \mathcal{L}_{focal}\left(M, \hat{M}\right).
\end{aligned}
\label{eq:segmentation loss}
\end{equation}
Where $M$ is the ground truth mask of the input image and $\hat{M}$ is the output of the segmentation sub-network. Inspired by ~\cite{zhang2023prn}, smooth L1 loss~\cite{girshick2015fast-rcnn} and focal loss~\cite{lin2017focal} are applied simultaneously to reduce over-sensitivity to outliers and accurately segment hard anomalous examples. $\gamma\in \mathbb{R}^{+}$ is a hyperparameter that controls the importance of $\mathcal{L}_{focal}$. Thus, the total loss used in jointly training DiffusionAD is:
\begin{equation}
\begin{aligned}
\mathcal{L}_{total}=\mathcal{L}_{noise} + \mathcal{L}_{mask}.
\end{aligned}
\label{eq:total loss}
\end{equation}

\noindent\textbf{Inference stage.\quad}Unlike other diffusion methods that reconstruct images in an iterative manner, we still perform a one-step norm-guided estimation in the inference stage, which is hundreds of times faster while maintaining comparable sampling quality. 
The robust decision boundary learned by the segmentation network will also effectively mitigate the effect of sub-optimal reconstruction quality. 
Moreover, we empirically set $t_s = 200$ and $t_b =400$ in the inference phase. 
After the segmentation model predicts the pixel-level anomaly score $\hat{M}$, we take the average of the top $K$ scores in $\hat{M}$ as the image-level anomaly score~\cite{zhang2023prn}.

\subsection{Anomaly Synthetic Strategy}
\label{subsec:Anomaly Synthetic Strategy}
Since prior information about anomalies is not available for training, we synthesize pseudo-anomalies online for end-to-end training. The idea of our anomaly synthesis strategy is adding visually inconsistent appearances to the normal samples inspired by ~\cite{zhang2023prn,zavrtanik2021draem,yang2023memseg}, and these out-of-distribution regions are defined as the synthesized anomalous regions.

\begin{figure}[h]
  \centering
  \includegraphics[width=1.0\linewidth]{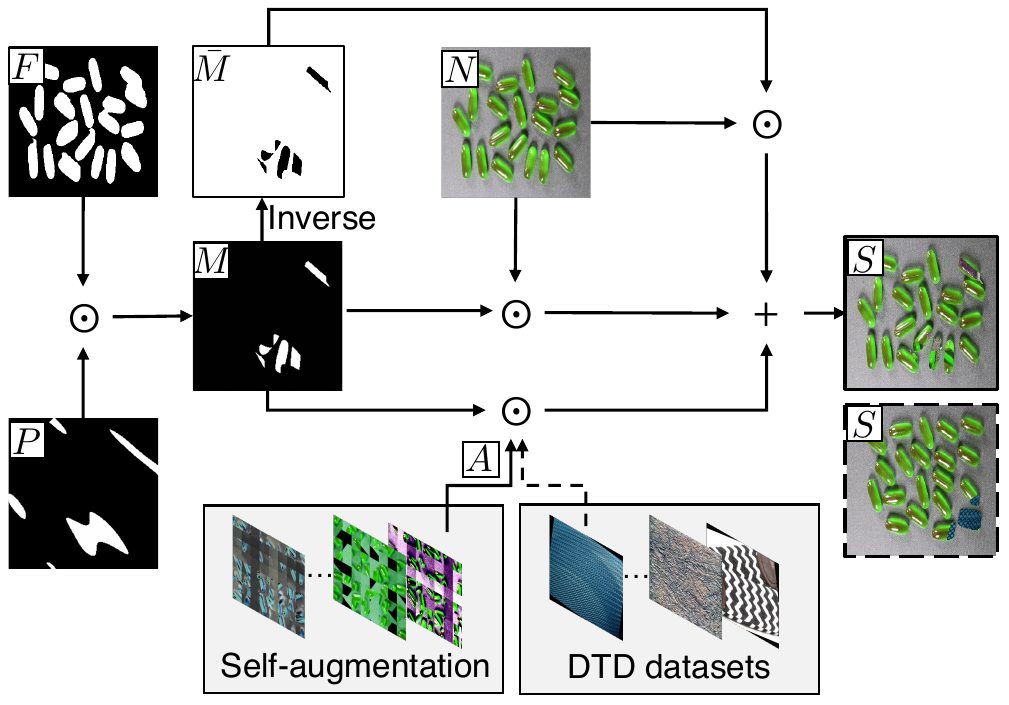}
  \caption{Visually inconsistent appearances are added to the normal samples to obtain synthetic anomalies.}
  \label{fig:Anomaly synthetic strategy}
\end{figure} 

Figure \ref{fig:Anomaly synthetic strategy} illustrates the overall process of transforming a normal sample (Figure \ref{fig:Anomaly synthetic strategy}, $N$) into a synthetic anomalous sample (Figure \ref{fig:Anomaly synthetic strategy}, $S$). Random and irregular anomalous regions (Figure \ref{fig:Anomaly synthetic strategy}, $P$) are first obtained from the Perlin~\cite{perlin1985perlin} noise and then multiplied by the object foreground~\cite{qin2022foreground} (Figure \ref{fig:Anomaly synthetic strategy}, $F$) of the normal sample to obtain the ground truth mask (Figure \ref{fig:Anomaly synthetic strategy}, $M$). For textural datasets, the foreground is replaced by a random part of the whole image. The appearance of visual inconsistencies (Figure \ref{fig:Anomaly synthetic strategy}, $A$) mainly stems from the self-augmentation of normal samples or Describing Textures Dataset (DTD)~\cite{cimpoi2014DTD}. The proposed synthetic anomaly (Figure \ref{fig:Anomaly synthetic strategy}, $S$) is defined as: 
\begin{equation}
\begin{aligned}
S=\beta(M \odot N)+(1-\beta)(M \odot A)+\bar{M} \odot N
\end{aligned}
\label{eq:anomaly synthetic}
\end{equation}
where $\bar{M}$ is the pixel-wise inverse operation of $M$, $\odot$ is the element-wise multiplication operation, and $\beta$ serves as an opacity parameter designed to enhance the fusion of anomalous and normal regions.

\section{Experiments}
\label{sec:experiments}

\subsection{Experimental Details}
\label{subsec:Experimental Details}

\noindent\textbf{Datasets.\quad}To assess the efficacy and generalizability of our approach, we conduct experiments on four diverse unsupervised datasets including MVTec~\cite{bergmann2019mvtec}, VisA~\cite{zou2022spd}, DAGM~\cite{wieler2007dagm}, and MPDD~\cite{jezek2021mpdd}.
These datasets contain samples of many different types of surface defects, such as scratches, cracks, holes, and depressions. 
The MVTec dataset is widely used to evaluate and test anomaly detection algorithms and industrial defect detection, consisting of 4,096 normal and 1,258 abnormal images. It contains samples of many different types of surface defects, such as scratches, cracks, holes, and depressions. 
VisA is a novel and challenging dataset consisting of 10,821 images (9,621 normal and 1,200 abnormal images) in total, which is $2 \times$ larger than MVTec. 
Visa contains 12 subsets, which can be divided into three broad categories based on the properties of the objects. The first category consists of four printed circuit boards (PCBS) with complex structures. The second type is a dataset with multiple instances in one view and consists of Capsules, Candles, Macaroni1, and Marcaroni2. The third type is single instances with roughly aligned objects: Cashew, Chewing gum, Fryum, and Pipe fryum. 
Anomalous images contain a variety of imperfections, including surface defects such as scratches, dents, colored spots, or cracks, and structural defects such as misplacement or missing parts.
DAGM consists of ten texture classes with 15,000 normal images and 2,100 abnormal images. Various defects that are visually close to the background, such as scratches and specks, constitute anomalous samples. 
MPDD contains 1064 normal images and 282 abnormal images. It focuses on metal fabrication and reflects real-world situations encountered on manually operated production lines.

\begin{table*}[ht]
\centering
\caption{Anomaly Detection and Localization on VisA. The best results of Image AUROC / Pixel PRO are highlighted in bold.}
\begin{tabular}{@{}lcccccccc@{}}
\toprule
                           & \multicolumn{4}{c}{Feature embedding-based} & \multicolumn{4}{c}{Generative model-based}                      \\ \cmidrule(l){2-5}  \cmidrule(l){6-9}
\multirow{-2}{*}{Category} & PatchCore~\cite{roth2022patchcore}   & RD4AD~\cite{deng2022rd4ad}      & RD++~\cite{tien2023rd++}       & SimpleNet~\cite{liu2023simplenet}  & DMAD~\cite{liu2023dmad}  & DRAEM~\cite{zavrtanik2021draem}     & FastFlow~\cite{yu2021fastflow}  & \textbf{Ours}      \\ \midrule
candle                     & 98.6/94.0     & 92.2/92.2  & 96.4/93.8   & \textbf{98.7}/89.0    & 92.7/90.6                   & 94.4/93.7 & 92.8/86.7 & \textbf{98.7/97.0}   \\
capsules                   & 81.6/85.5   & 90.1/56.9    & 92.1/95.8   & 89.9/91.4  & 88.0/88.4                     & 76.3/84.5 & 71.2/33.3 & \textbf{97.9/98.7} \\
cashew                     & 97.3/\textbf{94.5}   & \textbf{99.6}/79.0    & 97.8/91.2   & 97.5/82.8  & 95.0/88.8                     & 90.7/51.8 & 91.0/68.3   & 96.5/91.8 \\
chewinggum                 & 99.1/84.6   & 99.7/92.5    & 96.4/88.1   & 99.8/85.3  & 97.4/73.9                   & 94.2/60.4 & 91.4/74.8 & \textbf{99.9/92.2} \\
fryum                      & 96.2/85.3   & 96.6/81.0    & 95.8/90.0    & 98.1/87.8  & 98.0/92.2                     & 97.4/93.1 & 88.6/74.3 & \textbf{98.3/96.5} \\
macaroni1                  & 97.5/95.4   & 98.4/71.3    & 94.0/96.9    & 99.4/\textbf{98.9}  & 94.3/97.1                   & 95.0/96.7   & 98.3/77.7 & \textbf{99.5}/98.5 \\
macaroni2                  & 78.1/94.4   & 97.6/68.0    & 88.0/97.7    & 82.4/97.3  & 90.4/98.5                   & 96.2/\textbf{99.6} & 86.3/43.4 & \textbf{99.0/99.6}   \\
pcb1                       & 98.5/94.3   & 97.6/43.2    & 97.0/95.8    & 99.0/91.1    & 95.8/96.2                   & 54.8/24.8 & 77.4/59.9 & \textbf{99.2/96.9} \\
pcb2                       & 97.3/89.2   & 91.1/46.4    & 97.2/90.6  & \textbf{99.1}/91.0    & 96.9/89.3                   & 77.8/49.4 & 61.9/40.7 & \textbf{99.1/94.2} \\
pcb3                       & 97.9/90.9   & 95.5/80.3    & 96.8/93.1  & 98.5/93.0    & 98.3/93.6                   & 94.5/89.7 & 74.3/61.5 & \textbf{98.6/94.9} \\
pcb4                       & 99.6/90.1   & 96.5/72.2    & \textbf{99.8}/91.9  & 99.6/64.5  & 99.7/91.4                   & 93.4/64.3 & 80.9/58.8 & 98.9/\textbf{94.6} \\
pipe\_fryum                & \textbf{99.8}/95.7   & 97.0/68.3    & 99.6/95.6  & 99.7/91.7  & 99.0/95.3                     & 99.4/75.9 & 72.0/38.0     & \textbf{99.8/97.2} \\ \midrule
Average                    & 95.1/91.2   & 96.0/70.9    & 95.9/93.4  & 96.8/88.7  & 95.5/91.3                   & 88.7/73.7 & 82.2/59.8 & \textbf{98.8/96.0}   \\ \bottomrule
\end{tabular}

\label{tab:visa image auroc and pixel pro}
\end{table*}

\noindent\textbf{Evaluation Metrics.\quad}To measure the performance of anomaly detection (image-level), we report the Area Under Receiver Operator Characteristic curve (Image AUROC), the most widely used metric.
In terms of anomaly localization (pixel level) performance, we report three metrics: Pixel AUROC, PRO, and Pixel AP.
Pixel AUROC may provide an inflated view of the performance~\cite{zou2022spd}, which may pose challenges in measuring the true capabilities of the model when using only this metric.
The false positive rate is dominated by the extremely high number of non-anomalous pixels and remains low despite the false positive prediction~\cite{tao2022al_survey}, which is caused by the fact that anomalous regions typically only occupy a tiny fraction of the entire image. 
Thus, to comprehensively evaluate the anomaly localization performance, Per Region Overlap (PRO) and pixel-level Average Precision (AP) also play the role of evaluation metrics.
Per Region Overlap (PRO) metric is more capable of assessing the ability of fine-grained anomaly localization, which treats anomaly regions of varying sizes equally and is widely employed by previous works~\cite{deng2022rd4ad,roth2022patchcore,zhang2023prn}.
AP is more appropriate for highly imbalanced classes, especially for industrial anomaly localization, where accuracy is critical.

\noindent\textbf{Implementation Details.\quad}All images in the four datasets are resized to $256 \times 256$.
For the denoising sub-network, we adopt the UNet~\cite{ronneberger2015U-Net} architecture for estimating $\epsilon_{\theta}$. This architecture is mainly based on PixelCNN~\cite{salimanspixelcnn++} and Wide ResNet~\cite{he2016resnet-18}, with sinusoidal positional embedding~\cite{vaswani2017attention} to encode the time step $t$. $T$ is set to 1000 and is divided by $\tau = 300$ into two parts. The noise schedule is set to linear. We set the base channel to 128, the attention resolutions to 32, 16, 8, and the number of heads to 4. We do not use  EMA. For the segmentation sub-network, we employ a U-Net-like architecture consisting of an encoder, a decoder, and skip connections. $\gamma$ in the segmentation loss $\mathcal{L}_{seg}$ is set to 5. We train for 3000 epochs with a batch size of 16 consisting of 8 normal samples and 8 anomalous synthetic samples. We use Adam optimizer~\cite{kingma2014adam} for optimization, with an initial learning rate $10^{-4}$. The image-level anomaly score is obtained by taking the average of the top 50 anomalous pixels in pixel-wise anomaly score $\hat{M}$. We implement our model and experiments on NVIDIA A100 GPUs.

\subsection{Anomaly Detection and Localization Results}
We compare the anomaly detection and localization performance of DiffusionAD with four feature embedding-based methods, \ie, PatchCore~\cite{roth2022patchcore}, RD4AD~\cite{deng2022rd4ad}, RD++~\cite{tien2023rd++}, SimpleNet~\cite{liu2023simplenet} and three generative model-based methods, \ie, DMAD~\cite{liu2023dmad}, FastFlow~\cite{yu2021fastflow} and DRAEM~\cite{zavrtanik2021draem}.
\label{subsec:Anomaly Detection and Localization on four datasets}

\noindent\textbf{Per-class performance on VisA.\quad}The results of anomaly detection (Image AUROC) and anomaly localization (Pixel PRO) on the VisA dataset are shown in Table \ref{tab:visa image auroc and pixel pro}. 
Our method achieves the highest image AUROC and the highest PRO in 10 out of 12 classes. 
The average image AUROC results show that our method outperforms feature embedding-based SOTA by 2.0\% and generative model-based SOTA by 3.3\%. 
Meanwhile, for PRO, our method outperforms feature embedding-based SOTA by 2.6\% and generative model-based SOTA by 4.7\%.
In some hard cases, such as capsules and macaroni2, DiffusionAD outperforms previous generative model-based methods by a large margin.
In addition, DiffusionAD outperforms the previous generative model-based SOTA by 4.8\% on average pixel AP, as shown in Table \ref{tab: visa pixel auroc and pixel ap}.

\begin{table*}[h]
\centering
\caption{Anomaly Localization on VisA~\cite{bergmann2019mvtec}. The best results of Pixel AUROC / Pixel AP are highlighted in bold.}
\begin{tabular}{@{}lcccccccc@{}}
\toprule
                           & \multicolumn{4}{c}{Feature embedding-based} & \multicolumn{4}{c}{Generative model-based}                      \\ \cmidrule(l){2-5}  \cmidrule(l){6-9}
\multirow{-2}{*}{Category} & PatchCore~\cite{roth2022patchcore}   & RD4AD~\cite{deng2022rd4ad}      & RD++~\cite{tien2023rd++}       & SimpleNet~\cite{liu2023simplenet}  & DMAD~\cite{liu2023dmad}  & DRAEM~\cite{zavrtanik2021draem}     & FastFlow~\cite{yu2021fastflow}  & \textbf{Ours}      \\ \midrule
candle      & \textbf{99.5}/17.6 & 97.9/19.0   & 98.6/16.3 & 97.7/12.8 & 98.1/18.9 & 97.3/32.7 & 94.9/4.9  & 99.2/\textbf{49.7} \\
capsules    & 99.5/\textbf{68.7} & 89.5/12.3 & 99.4/57.1 & 99.0/57.9   & 99.2/50.5 & 99.1/47.9 & 75.3/1.4  & \textbf{99.5}/51.8 \\
cashew      & 98.9/58.5 & 95.8/28.2 & 95.8/53.9 & 98.8/58.6 & 95.3/57.3 & 88.2/28.5 & 91.4/9.2  & \textbf{98.9/61.4} \\
chewinggum  & 99.1/42.8 & 99.0/58.9   & \textbf{99.4/65.8} & 98.3/17.7 & 97.9/56.1 & 97.1/39.0   & 98.6/43.6 & 99.3/50.0   \\
fryum       & 93.8/37.2 & 94.3/54.3 & 96.5/49.7 & 91.1/38.0   & 97.0/49.2   & 92.7/41.1 & 97.3/35.3 & \textbf{98.2/51.5} \\
macaroni1   & \textbf{99.8}/7.6  & 97.7/43.8 & 99.7/21.1 & 99.6/6.7  & 99.7/16.3 & 99.7/\textbf{45.8} & 97.3/27.8 & 99.2/41.3 \\
macaroni2   & 99.1/3.2  & 87.7/35.2 & 99.7/11.1 & 98.9/4.4  & 99.7/9.8  & 99.9/41.0   & 89.2/29.8 & \textbf{99.9/44.1} \\
pcb1        & \textbf{99.9/91.7} & 75.0/0.5  & 99.7/82.6 & 99.6/88.6 & 99.8/82.8 & 90.5/28.1 & 75.2/0.2  & 97.8/25.6 \\
pcb2        & 99.0/14.3 & 64.8/0.6  & 99.0/14.8   & 98.3/12.5 & \textbf{99.0/14.7}   & 90.5/4.9  & 67.3/0.1  & 98.6/15.0   \\
pcb3        & 99.2/38.9 & 95.5/59.1 & 99.2/20.1 & 99.2/44.3 & 99.3/29.6 & 98.6/20.3 & 94.8/23.7 & \textbf{99.4/50.6} \\
pcb4        & 98.6/42.4 & 92.8/10.6 & 98.6/42.6 & 93.9/34.0   & \textbf{98.8/46.8} & 88.0/20.6   & 89.9/5.4  & 97.6/43.1 \\
pipe\_fryum & 99.1/58.5 & 92.0/10.4   & 99.1/54.9 & 98.9/60.5 & \textbf{99.3}/60.2 & 90.9/16.5 & 87.3/5.4  & 99.0/\textbf{64.9}   \\ \midrule
Average     & 98.8/40.1 & 90.1/27.7 & 98.7/40.8 & 97.8/36.3 & 98.6/41.0   & 94.4/30.5 & 88.2/15.6 & \textbf{98.9/45.8} \\ \bottomrule
\end{tabular}

\label{tab: visa pixel auroc and pixel ap}
\end{table*}

\noindent\textbf{Per-class performance on MVTec.\quad}Table \ref{tab: mvtec image auroc and pixel auroc} and Table \ref{tab: mvtec pixel pro and pixel ap} show the comparison of anomaly detection and localization performance on the MVTec~\cite{bergmann2019mvtec} dataset, indicating that DiffusionAD significantly surpasses previous generative model-based SOTA in anomaly localization, especially with improvements of 3.6\% and 7.7\% on pixel PRO and AP, respectively.

\begin{table*}[]
\centering
\caption{Anomaly Detection and Localization on MVTec~\cite{bergmann2019mvtec}. The best results of Image AUROC / Pixel AUROC are highlighted in bold.}
\begin{tabular}{@{}lcccccccc@{}}
\toprule
                           & \multicolumn{4}{c}{Feature embedding-based} & \multicolumn{4}{c}{Generative model-based}                      \\ \cmidrule(l){2-5}  \cmidrule(l){6-9}
\multirow{-2}{*}{Category} & PatchCore~\cite{roth2022patchcore}   & RD4AD~\cite{deng2022rd4ad}      & RD++~\cite{tien2023rd++}       & SimpleNet~\cite{liu2023simplenet}  & DMAD~\cite{liu2023dmad}  & DRAEM~\cite{zavrtanik2021draem}     & FastFlow~\cite{yu2021fastflow}  & \textbf{Ours}      \\ \midrule
Carpet     & 98.7/99.0   & 98.9/98.8 & \textbf{100/99.2}  & 99.0/98.4   & \textbf{100}/99.1  & 97.0/95.5   & 87.5/98.0   & 99.3/99.0   \\
Grid       & 98.2/98.7 & \textbf{100}/97.0    & \textbf{100}/99.3  & 97.8/98.9 & \textbf{100}/99.2  & 99.9/\textbf{99.7} & 88.1/93.5 & \textbf{100/99.7}  \\
Leather    & \textbf{100}/99.3  & \textbf{100}/98.6  & \textbf{100}/99.4  & 99.3/98.0   & \textbf{100}/99.5  & \textbf{100}/98.6  & 96.1/93.0   & \textbf{100/99.8}  \\
Tile       & 98.7/95.6 & 99.3/98.9 & 99.7/96.6 & 99.7/96.6 & \textbf{100}/96.0    & 99.6/99.2 & 63.2/96.1 & 99.9/\textbf{99.7} \\
Wood       & 99.2/95.0   & 99.2/\textbf{99.3} & 99.3/95.8 & 99.9/93.2 & \textbf{100}/95.5  & 99.1/96.4 & 90.8/95.9 & 99.9/96.7 \\
Bottle     & \textbf{100}/98.6  & \textbf{100}/99.0    & \textbf{100}/98.8  & \textbf{100}/94.0    & \textbf{100}/98.9  & 99.2/\textbf{99.1} & \textbf{100}/92.7  & 99.7/\textbf{99.1} \\
Cable      & 99.5/98.4 & 95.0/\textbf{99.4}   & 99.2/98.4 & \textbf{99.9}/98.7 & 99.1/98.1 & 91.8/94.7 & 91.2/97.5 & 99.6/98.1 \\
Capsule    & 98.1/\textbf{98.8} & 96.3/97.3 & 99.0/\textbf{98.8}   & \textbf{100}/97.4  & 98.9/98.3 & 98.5/94.3 & 99.2/97.4 & 98.9/98.3  \\
Hazelnut   & \textbf{100}/98.7  & 99.9/98.2 & \textbf{100}/99.2  & \textbf{100}/98.6  & \textbf{100}/99.1  & \textbf{100/99.7}  & 98.8/98.9 & \textbf{100/99.7} \\
Metal nut  & \textbf{100}/98.4  & \textbf{100}/99.6  & \textbf{100}/98.1  & \textbf{100}/97.0    & \textbf{100}/97.7  & 98.7/99.5 & 90.8/98.1 & \textbf{100/99.7}  \\
Pill       & 96.6/97.4 & 99.6/95.7 & 98.4/98.3 & \textbf{100}/98.8  & 97.3/98.7 & 98.9/97.6 & 98.0/92.3   & 99.0/\textbf{99.3}   \\
Screw      & 98.1/99.4 & 97.0/99.1   & 98.9/\textbf{99.7} & \textbf{100}/98.0    & \textbf{100}/99.6  & 93.9/97.6 & 97.8/99.4 & 99.4/99.0   \\
Toothbrush & \textbf{100}/98.7  & 99.5/93.0   & \textbf{100}/99.1  & 98.1/99.2 & \textbf{100/99.4}  & \textbf{100}/98.1  & \textbf{100}/92.7  & \textbf{100}/98.8  \\
Transistor & \textbf{100}/96.3  & 96.7/95.4 & 98.5/94.3 & \textbf{100/97.8}  & 98.7/95.4 & 93.1/90.9 & 88.1/94.3 & 99.8/92.9 \\
Zipper     & 99.4/98.8 & 98.5/98.2 & 98.6/98.8 & \textbf{100}/99.2  & 99.6/98.3 & \textbf{100}/98.8  & 67.5/93.4 & \textbf{100/99.2}  \\ \midrule
Average    & 99.1/98.1 & 98.5/97.8 & 99.4/98.3 & 99.6/97.6 & 99.6/98.2 & 98.0/97.3   & 90.5/95.5 & \textbf{99.7/98.7} \\ \bottomrule
\end{tabular}

\label{tab: mvtec image auroc and pixel auroc}
\end{table*}

\begin{table*}[]
\centering
\caption{Anomaly Localization on MVTec~\cite{bergmann2019mvtec}. The best results of Pixel PRO / Pixel AP are highlighted in bold.}
\begin{tabular}{@{}lcccccccc@{}}
\toprule
                           & \multicolumn{4}{c}{Feature embedding-based} & \multicolumn{4}{c}{Generative model-based}                      \\ \cmidrule(l){2-5}  \cmidrule(l){6-9}
\multirow{-2}{*}{Category} & PatchCore~\cite{roth2022patchcore}   & RD4AD~\cite{deng2022rd4ad}      & RD++~\cite{tien2023rd++}       & SimpleNet~\cite{liu2023simplenet}  & DMAD~\cite{liu2023dmad}  & DRAEM~\cite{zavrtanik2021draem}     & FastFlow~\cite{yu2021fastflow}  & \textbf{Ours}      \\ \midrule
Carpet     & 96.6/76.8 & 97.0/78.7   & \textbf{97.7}/63.6 & 93.2/\textbf{79.8} & 86.1/44.6 & 92.0/53.5   & 84.2/68.8 & 95.0/77.2   \\
Grid       & 95.9/65.3 & 97.6/52.7 & 97.7/48.5 & 92.8/42.1 & 72.4/7.4  & 97.8/65.7 & 83.8/27.6 & \textbf{98.9/68.8} \\
Leather    & 98.9/44.2 & 99.1/45.1 & \textbf{99.2}/50.1 & 87.4/43.3 & 97.7/41.3 & 96.8/75.3 & 91.0/41.8   & 98.9/\textbf{69.2} \\
Tile       & 87.4/62.7 & 90.6/57.4 & 92.4/53.4 & 87.4/28.8 & 82.7/54.3 & 97.4/92.3 & 81.8/20.5 & \textbf{98.3/97.1} \\
Wood       & 89.6/32.5 & 90.9/49.2 & \textbf{93.3}/52.0   & 86.2/59.8 & 86.3/44.4 & 92.8/\textbf{77.7} & 85.1/53.9 & 90.3/77.4 \\
Bottle     & 96.1/53.7 & 96.6/62.1 & \textbf{97.0}/79.9   & 82.2/45.2 & 96.0/78.9   & 96.4/86.5 & 91.9/45.5 & 96.8/\textbf{89.7} \\
Cable      & 92.6/45.6 & 91.0/48.2   & 93.9/62.5 & 94.7/63.7 & 95.1/58.3 & 75.4/52.4 & 80.8/35.1 & \textbf{95.5/72.5} \\
Capsule    & 95.5/\textbf{87.0}   & \textbf{95.8}/79.1 & 96.4/48.1 & 92.0/66.6   & 89.7/41.4 & 90.4/49.4 & 76.7/27.6 & 93.3/59.2 \\
Hazelnut   & 93.9/77.7 & 95.5/78.4 & 96.3/65.4 & 91.9/40.4 & 96.4/60.2 & 97.5/92.9 & 89.0/63.3   & \textbf{97.5/92.6} \\
Metal nut  & 91.3/35.4 & 92.3/53.6 & 93.0/82.3   & 93.1/68.1 & 94.6/83.5 & 93.2/96.3 & 86.6/31.1 & \textbf{97.8/97.1} \\
Pill       & 94.1/54.6 & 96.4/53.2 & 97.0/78.1   & 86.4/92.0   & 95.0/77.6   & 88.1/48.5 & 94.6/25.9 & \textbf{98.7/86.9} \\
Screw      & 97.9/37.2 & 98.2/51.8 & \textbf{98.6}/54.4 & 89.9/\textbf{70.2} & 94.3/48.2 & 97.0/58.2   & 91.5/52.5 & 91.6/59.4 \\
Toothbrush & 91.4/61.0   & 94.5/54.9 & 94.2/53.8 & 95.4/33.6 & 92.5/52.5 & 89.9/44.7 & \textbf{98.0}/60.9   & 93.7/\textbf{61.1} \\
Transistor & 83.5/47.7 & 78.0/48.2   & 81.8/58.6 & 81.2/46.5 & 86.9/\textbf{60.7} & 81.0/50.7   & 73.1/37.8 & \textbf{91.7}/58.2 \\
Zipper     & \textbf{97.1}/59.5 & 95.4/57.1 & 96.3/60.9 & 96.6/42.6 & 92.9/49.6 & 95.4/\textbf{81.5} & 75.6/4.5  & 96.5/74.2 \\\midrule
Average    & 93.5/56.1 & 93.9/58.0   & 95.0/60.8   & 90.0/54.8   & 90.6/53.5 & 92.1/68.4 & 85.6/39.8 & \textbf{95.7/76.1} \\ \bottomrule
\end{tabular}
\label{tab: mvtec pixel pro and pixel ap}
\end{table*}

\begin{table*}[]
\centering
\caption{Comparison of DiffusionAD with other approaches on four datasets. ``I'', ``P'', ``O'' and ``A'' respectively refer to the four metrics of Image AUROC, Pixel AUROC, Pixel PRO, and Pixel AP. The best results are highlighted in bold.}
\scalebox{0.9}{
\begin{tabular}{@{}lcccccccccccccccccccc@{}}
\toprule
\multirow{2}{*}{} & \multicolumn{4}{c}{VisA~\cite{zou2022spd}}  & \multicolumn{4}{c}{MVTec~\cite{bergmann2019mvtec}} & \multicolumn{4}{c}{DAGM~\cite{wieler2007dagm}}  & \multicolumn{4}{c}{MPDD~\cite{jezek2021mpdd}}  & \multicolumn{4}{c}{\textbf{Average}}\\ \cmidrule(lr){2-5} \cmidrule(lr){6-9} \cmidrule(lr){10-13} \cmidrule(lr){14-17} \cmidrule(lr){18-21} 
                  & I $\uparrow$    & P $\uparrow$    & O $\uparrow$    & A $\uparrow$    & I $\uparrow$    & P $\uparrow$    & O $\uparrow$    & A $\uparrow$    & I $\uparrow$    & P $\uparrow$    & O $\uparrow$    & A $\uparrow$    & I $\uparrow$    & P $\uparrow$    & O $\uparrow$    & A $\uparrow$  & I $\uparrow$    & P $\uparrow$    & O $\uparrow$    & A $\uparrow$  \\ \midrule
PatchCore    & 95.1 & 98.8 & 91.2 & \multicolumn{1}{c|}{40.1} & 99.1 & 98.1 & 93.5 & \multicolumn{1}{c|}{56.1} & 93.6 & 96.7 & 89.3 & \multicolumn{1}{c|}{51.7} & 94.8 & \textbf{99.0}   & 93.9 & \multicolumn{1}{c|}{43.2} 
& 95.7 & 98.2   & 92.0 & 47.8\\ 
RD4AD        & 96.0 & 90.1 & 70.9 & \multicolumn{1}{c|}{27.7} & 98.5 & 97.8 & 93.9 & \multicolumn{1}{c|}{58.0} & 95.8 & \textbf{97.5} & 93.0 & \multicolumn{1}{c|}{53.4} & 92.7 & 98.7 & \textbf{95.3} & \multicolumn{1}{c|}{45.5}
& 95.8 & 96.0 & 88.3 & 46.2\\ 
RD++         & 95.9 & 98.7 & 93.4 & \multicolumn{1}{c|}{40.8} & 99.4 & 98.3 & 95.0 & \multicolumn{1}{c|}{60.8} & 98.5 & 97.4 & \textbf{93.8} & \multicolumn{1}{c|}{64.3} & 92.9 & 98.3 & 94.9 & \multicolumn{1}{c|}{43.7} 
&96.7	&98.2	&94.3	&52.4\\ 
SimpleNet    & 96.8 & 97.8 & 88.7 & \multicolumn{1}{c|}{36.3} & 99.6 & 97.6 & 90.0   & \multicolumn{1}{c|}{54.8} & 95.3 & 97.1 & 91.3 & \multicolumn{1}{c|}{48.1} & 96.1 & 97.6 & 91.2 & \multicolumn{1}{c|}{40.7}
&96.9	&97.5	&90.3	&45.0\\ \midrule 
DMAD         & 95.5 & 98.6 & 91.3 & \multicolumn{1}{c|}{41.0} & 99.6 & 98.2 & 90.6 & \multicolumn{1}{c|}{53.5} & 89.1 & 92.5 & 83.8 & \multicolumn{1}{c|}{46.0} & 91.7 & 98.0   & 92.6 & \multicolumn{1}{c|}{42.6}
&94.0	&96.8	&89.6	&45.8\\ 
FastFlow     & 82.2 & 88.2 & 59.8 & \multicolumn{1}{c|}{15.6} & 90.5 & 95.5 & 85.6 & \multicolumn{1}{c|}{39.8} & 87.4 & 91.1 & 79.9 & \multicolumn{1}{c|}{34.2} & 88.7 & 80.8 & 49.8 & \multicolumn{1}{c|}{11.5}
&87.2	&88.9	&68.8	&25.3\\ 
DRAEM        & 88.7 & 94.4 & 73.7 & \multicolumn{1}{c|}{30.5} & 98.0 & 97.3 & 92.1 & \multicolumn{1}{c|}{68.4} & 90.8 & 86.8 & 71.0 & \multicolumn{1}{c|}{30.6} & 94.1 & 91.8 & 78.1 & \multicolumn{1}{c|}{28.8}
&92.9	&92.6	&78.7	&39.6\\ 
\rowcolor[HTML]{e9e9e9} 
\textbf{Ours}  & \textbf{98.8} & \textbf{98.9} & \textbf{96.0}   & \multicolumn{1}{c|}{\textbf{45.8}} & \textbf{99.7} & \textbf{98.7} & \textbf{95.7} & \multicolumn{1}{c|}{\textbf{76.1}} & \textbf{99.6} & \textbf{97.5} & \textbf{93.8} & \multicolumn{1}{c|}{\textbf{65.1}} & \textbf{96.2} & 98.5 & \textbf{95.3} & \multicolumn{1}{c|}{\textbf{45.8}}
&\textbf{98.6}	&\textbf{98.4}	&\textbf{95.2}	&\textbf{58.2}\\ \bottomrule 
\end{tabular}}
\label{tab:results of four datasets}
\end{table*}

\noindent\textbf{Quantitative results across the four datasets.\quad}Table \ref{tab:results of four datasets} enumerates the performance of the aforementioned methods across the four datasets. We conduct a comprehensive comparison based on four metrics, which include image AUROC for anomaly detection and pixel AUROC, PRO, and pixel AP for anomaly localization. 
In addition to its performance on the VisA dataset, DiffusionAD achieves state-of-the-art results on another widely used dataset, MVTec, across four key metrics. 
Remarkably, DiffusionAD outperforms the previous feature embedding-based SOTA by 15.3\% and the previous generative model-based SOTA by 7.7\% in terms of average pixel AP metric. 
Our method also outperforms previous approaches on the DAGM and MPDD datasets, especially in terms of average image AUROC.
Finally, we evaluate the generalization ability of the method by measuring its average performance across the four datasets. 
As shown in Table \ref{tab:results of four datasets}, our approach significantly outperforms previous methods and in particular, outperforms previous generative models by a large margin. 

\begin{table}[h]
\centering
\caption{Comparison with other diffusion-based methods.}
\begin{tabular}{@{}lcccccc@{}}
\toprule
              & \multicolumn{2}{c}{Speed} & \multicolumn{4}{c}{Performance on MVTec} \\ \cmidrule(l){2-3} \cmidrule(l){4-7} 
              & Step $\downarrow$        & FPS $\uparrow$         & I $\uparrow$            & P $\uparrow$    & O $\uparrow$    & A $\uparrow$     \\ \midrule
DiffAD~\cite{zhang2023diffad}        & 50                  &  1.2        & 98.7   & 98.3   & N/A   & 74.6  \\
KLAD~\cite{lu2023KL_divergence} & 20          & N/A         & N/A    & 96.7   & 94.1  & N/A   \\\midrule
Ours          & \textbf{1}           & \textbf{23.5}        & \textbf{99.7}   & \textbf{98.7}   & \textbf{95.7}  & \textbf{76.1}  \\ \bottomrule
\end{tabular}
\label{tab:Comparison with other diffusion-based methods.}
\end{table}

\noindent\textbf{Comparison with other diffusion-based methods.\quad} We compare DiffusionAD with previous diffusion-based anomaly detection methods, DiffAD~\cite{zhang2023diffad} and KLAD~\cite{lu2023KL_divergence}, as shown in Table \ref{tab:Comparison with other diffusion-based methods.}. Compared to DiffAD and KLAD, which require 50 and 20 denoising steps, respectively, our proposed one-step norm-guided paradigm achieves a significantly faster FPS (\ie, 23.5 \vs 1.2). Moreover, our method outperforms them by a clear margin in various metrics, including Image AUROC, Pixel AUROC, Pixel PRO, and Pixel AP.

\begin{figure}[h]
  \centering
  \includegraphics[width=0.9\linewidth]{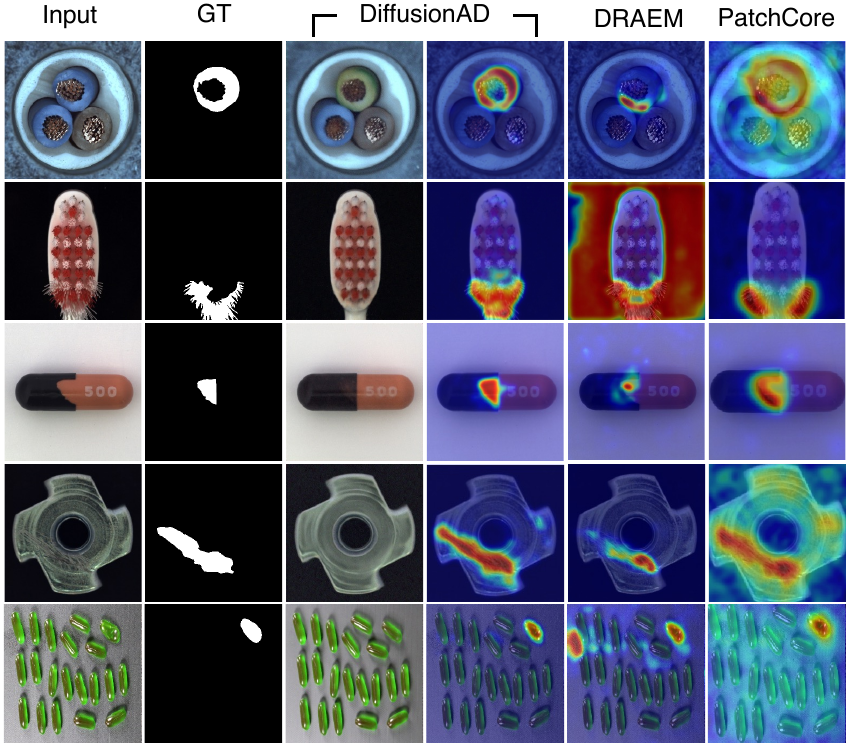}
  \caption{Qualitative results on MVTec~\cite{bergmann2019mvtec} and VisA~\cite{zou2022spd}. 
}
  \label{fig:visualization}
\end{figure}

\begin{figure}
  \centering
  \includegraphics[width=1.0\linewidth]{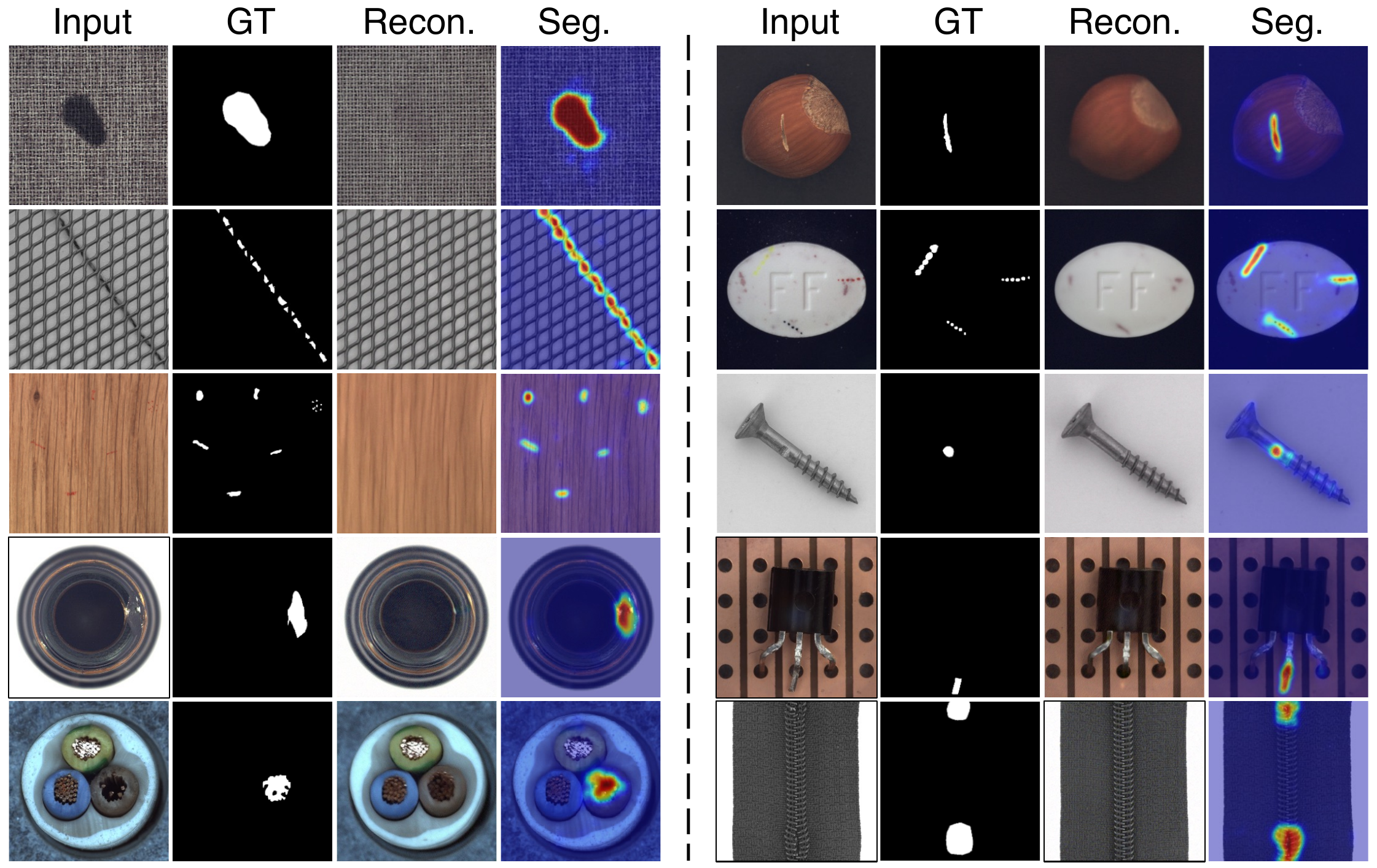}
  \caption{More qualitative examples on MVTec~\cite{bergmann2019mvtec}.}
  \label{fig: supplemental mvtec more visualization} 
\end{figure}

\begin{figure}
  \centering
  \includegraphics[width=1.0\linewidth]{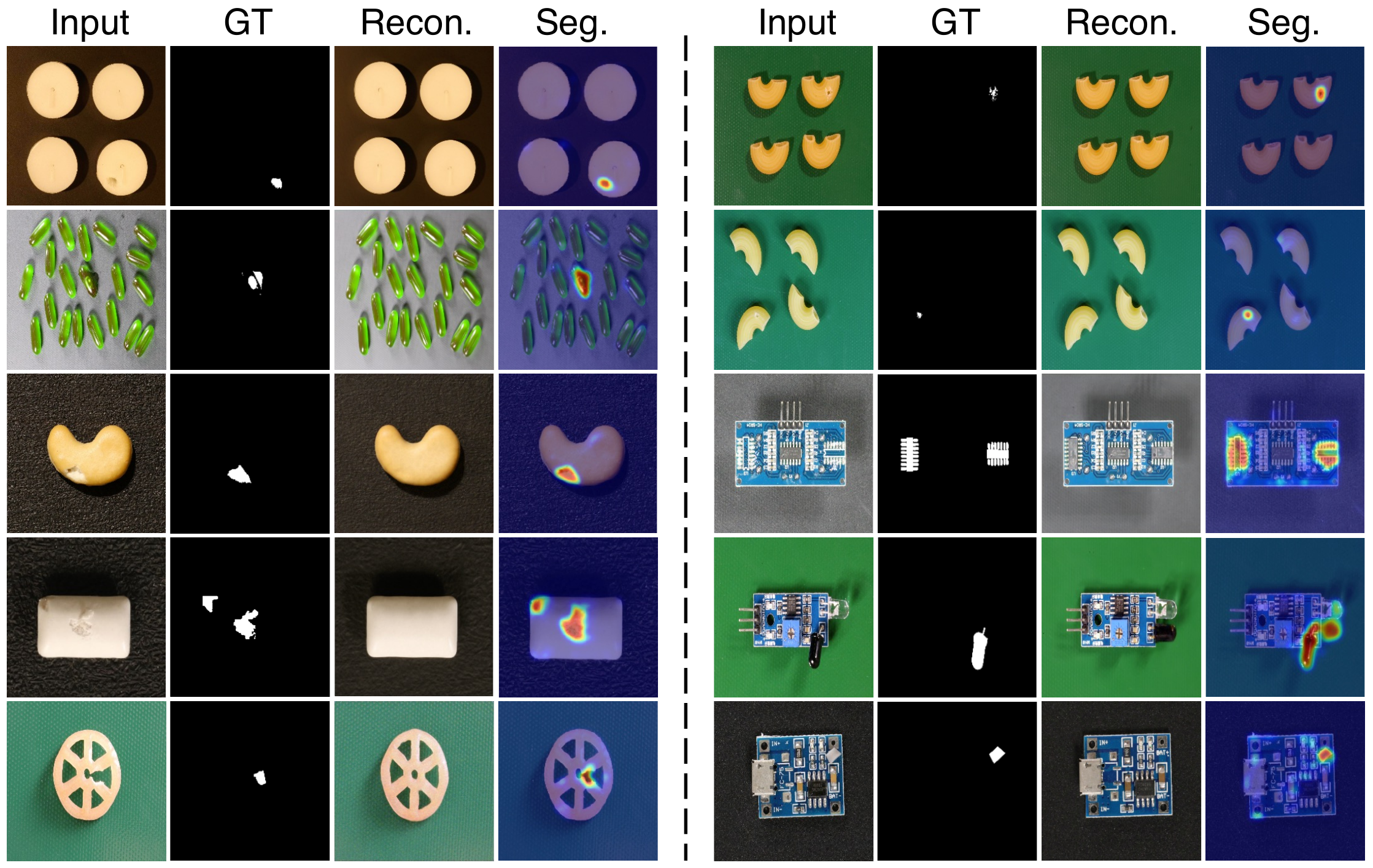}
  \caption{More qualitative examples on VisA~\cite{zou2022spd}.}
  \label{fig: supplemental visa more visualization} 
\end{figure}

\begin{figure}
  \centering
  \includegraphics[width=1.0\linewidth]{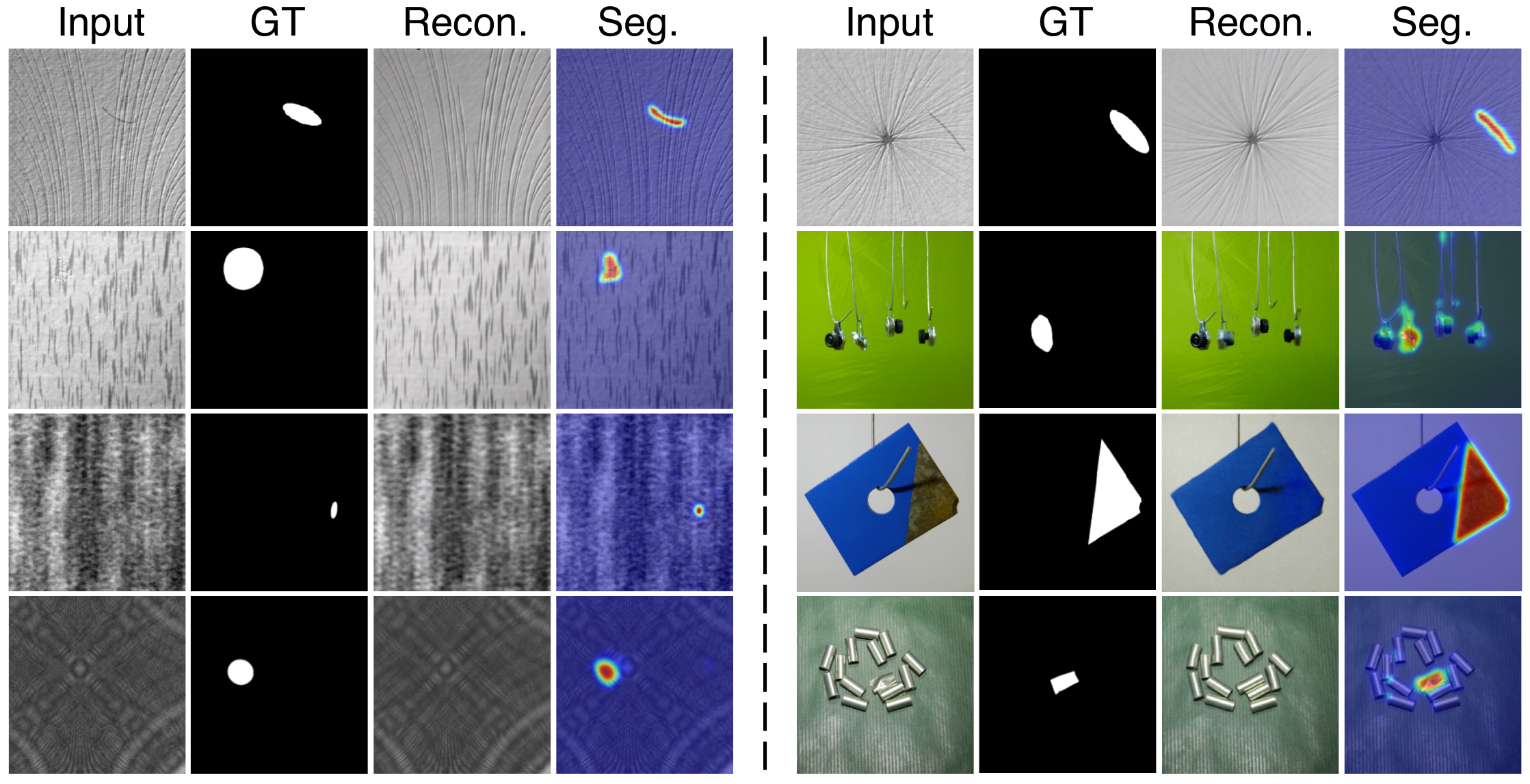}
  \caption{Qualitative examples on DAGM~\cite{wieler2007dagm} and MPDD~\cite{jezek2021mpdd}.}
  \label{fig: supplemental dagm and mpdd more visualization} 
\end{figure}

\noindent\textbf{Qualitatively evaluate on anomaly localization.\quad}According to the results in Figure \ref{fig:visualization}, we qualitatively assess the performance of DiffusionAD on anomaly detection and localization compared to the previous feature embedding-based method PatchCore~\cite{roth2022patchcore} and generative model-based method DRAEM~\cite{zavrtanik2021draem}. 
These images are sub-datasets of MVTec~\cite{bergmann2019mvtec} and Visa~\cite{zou2022spd}, where the anomalous regions vary in shape, size and number, as shown in the first and second columns of Figure \ref{fig:visualization}.
The third and fourth columns show the reconstruction and anomaly localization results of DiffusionAD, respectively. It can be observed that the reconstruction sub-network successfully repairs various types of anomalies while maintaining high image fidelity. After that, the segmentation sub-network accurately predicts the pixel-wise anomaly score using the input image and its anomaly-free reconstruction, leading to more accurate anomaly localization than previous methods such as DRAEM~\cite{zavrtanik2021draem} and PatchCore~\cite{roth2022patchcore}.
Figures \ref{fig: supplemental mvtec more visualization}, \ref{fig: supplemental visa more visualization}, and \ref{fig: supplemental dagm and mpdd more visualization} showcase additional detailed qualitative results of DiffusionAD on the four datasets.

\subsection{Ablation Study}
\label{subsec:Ablation Study}

\noindent\textbf{The importance of architecture.\quad}Table \ref{tab:ablation on module} illustrates the impact of different architectures and denoising methods on performance and inference speed. We introduce FPS (Frames Per Second) to evaluate the inference speed. As indicated in the first row of Table \ref{tab:ablation on module}, the performance of anomaly detection and localization is limited due to the unsatisfactory reconstruction results of the Autoencoder. However, its inference speed is notably fast. 
When we adopt the proposed noise-to-norm paradigm, as seen in the second row, which utilizes higher-quality reconstructions from the diffusion model, the performance is significantly enhanced. 

\noindent\textbf{The importance of denoising paradigm.\quad} In the second row of Table \ref{tab:ablation on module}, each denoising iteration corresponds to one network inference (in this case, 400 iterations in total), which falls significantly short of the real-world requirements for inference speed.
In the third row, we employ a faster sampler, DDIM~\cite{song2020ddim}, to reduce the denoising steps to 30. However, the FPS in the inference stage still lags significantly behind the previous diffusion-free methods~(\ie 0.92 FPS \vs 32.4 FPS).

In the fourth row, when we employ the proposed one-step denoising, the inference speed is approximately 300 times faster (26.7 FPS \vs 0.09 FPS)  than the iterative approach while achieving better anomaly detection and localization performance. 
We attribute this performance gain to two aspects: 
\Rmnum{1}) The reconstruction quality of one-step denoising is comparable to that of iterative methods.
\Rmnum{2}) Training-Inference consistency: During the end-to-end training stage, the segmentation sub-network is trained end-to-end to learn discrepancies between original inputs and their one-step denoised counterparts rather than iteratively denoised counterparts. Therefore, using iterative denoising during inference introduces distribution shifts between the segmentation sub-network's training inputs (one-step reconstructions) and inference inputs (iterative reconstructions), which reduces the accuracy of anomaly detection and localization.

In the last row, the norm-guided paradigm combines the advantages of one-step reconstructions from different noise scales, leading to further performance enhancements while maintaining comparable inference speed.

\begin{table*}[]
\centering
\caption{Impact of modules and denoising paradigms.}
\begin{tabular}{@{}cccccccccccc@{}}
\toprule
\multicolumn{3}{c}{Architecture} & \multicolumn{4}{c}{Denoising Paradigm } & \multicolumn{5}{c}{Performance}  \\ \cmidrule(lr){1-3} \cmidrule(lr){4-7} \cmidrule(lr){8-12}
Segmentation                & Auto-Encoder               & Diffusion Model              & Iterative-400   & DDIM-30  & One-step        & Norm-guided          & I $\uparrow$    & P $\uparrow$    & O $\uparrow$    & A $\uparrow$    & F $\uparrow$    \\ \midrule
$\checkmark$        & $\checkmark$      &                   &             &  &                     &               & 90.6 & 96.2 & 76.1 & 31.9 & \textbf{32.4} \\ 
$\checkmark$        &                   & $\checkmark$      & $\checkmark$ &                        &                 &     & 96.8 & 98.5 & 92.5   & 42.5 & 0.09 \\
$\checkmark$        &                   & $\checkmark$      &              & $\checkmark$ &         &                 &96.6 & 98.2 & 91.9 & 41.4 & 0.92 \\
$\checkmark$        &                   & $\checkmark$      &              &  & $\checkmark$        &                 & 98.0 & 98.6 & 94.7 & 44.2 & 26.7 \\
$\checkmark$        &                   & $\checkmark$      &              &  & $\checkmark$        & $\checkmark$   & \textbf{98.8} & \textbf{98.9} & \textbf{96.0}   & \textbf{45.8} & 23.5 \\ \bottomrule
\end{tabular}
\label{tab:ablation on module}
\end{table*}

\begin{figure}[h]
  \centering
  \includegraphics[width=1.0\linewidth]{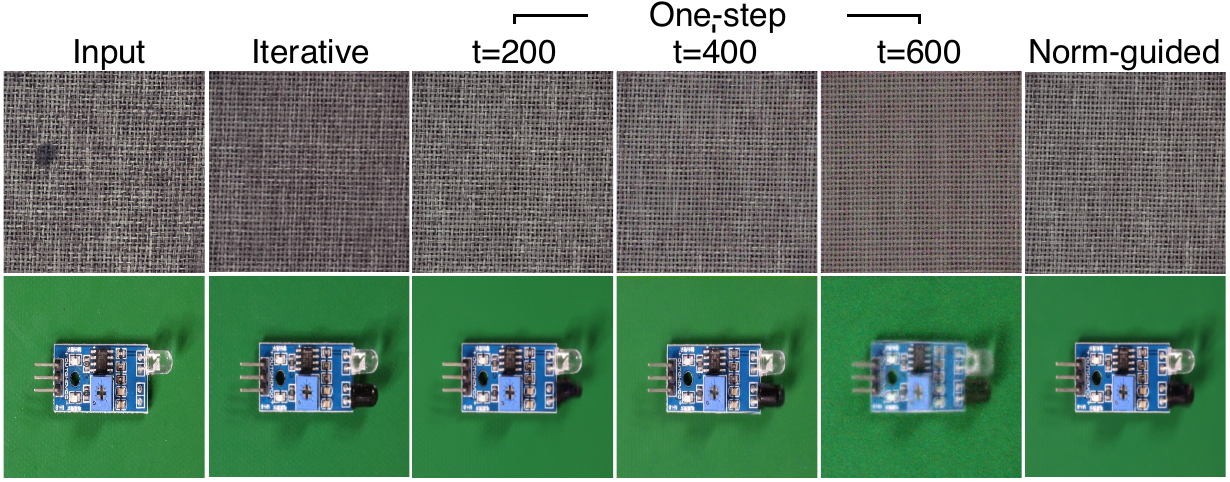}
  \caption{Impact of noise scale on anomaly-free reconstruction.}
  \label{fig:one step vs iterative recon}
\end{figure}

\noindent\textbf{The effect of noise scale.\quad}We investigate how the added noise scale of the input individually affects the reconstruction results and anomaly detection performance. 
First, Figure \ref{fig:one step vs iterative recon} illustrates the effect of the noise scale on the quality of the reconstruction.
  
The second column of the Figure \ref{fig:one step vs iterative recon} shows the iterative denoising results from $t = 400$, which can be considered as the upper bound of the reconstruction quality.
For one-step reconstruction, direct prediction becomes more challenging as the noise scale $t$ increases, resulting in lower pixel quality. However, repairing more pronounced anomalies often requires larger noise scales to perturb the anomalies, leading to reconstructions with higher semantic quality (the second row of Figure \ref{fig:one step vs iterative recon}).
Thus, the norm-guided one-step approach combines the advantages of both scales, resulting in a reconstruction that encompasses both pixel and semantic quality, closely approximating the quality of the iterative one.

\begin{figure}[h]
  \centering
  \includegraphics[width=0.95\linewidth]{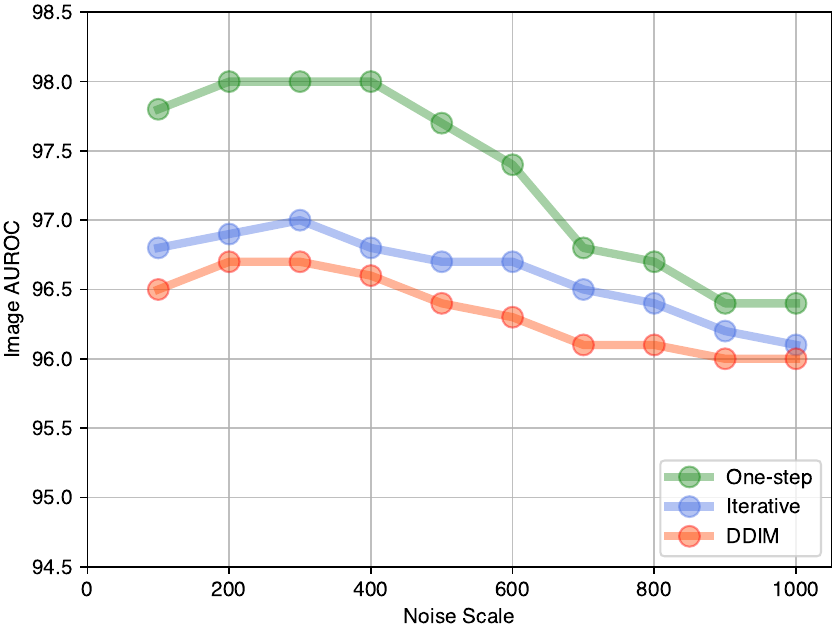}
  \caption{Impact of noise scale on anomaly detection performance.}
  \label{fig:one step vs iterative seg}
\end{figure}

Figure \ref{fig:one step vs iterative seg} further depicts the effect of different noise scales on the anomaly detection performance. 
When adopting the one-step denoising method, the inference time remains constant across different noise scales. However, there is a decrease in Image AUROC as $t$ becomes relatively large, \ie $t \geq 500$, mainly due to the reduced reconstruction quality.
When employing either the iterative denoising or DDIM denoising approach, there is little variation in the change of Image AUOROC as the noise scale increases. This is because these denoising paradigms yield high-quality reconstruction results at different noise scales.
However, the inference time of these paradigms increases with noise scales $t$.

\begin{table}[h]
\centering
\caption{Impact of different denoising paradigms on training and inference.}
\tabcolsep=0.10cm
\begin{tabular}{@{}cccccc@{}}
\toprule
\multicolumn{2}{c}{Train}   & \multicolumn{2}{c}{Inference} & \multicolumn{2}{c}{Performance} \\ \cmidrule(lr){1-2}\cmidrule(lr){3-4}\cmidrule(lr){5-6}
Paradigm  & Denoising steps & Paradigm   & Denoising steps  & I $\uparrow$                  & P $\uparrow$                  \\ \midrule
\rowcolor[HTML]{e9e9e9} 
\textbf{One-step}  & 1               & \textbf{One-step}   & 1                & 98.0                 & 98.6               \\
One-step  & 1               & Iterative  & 400              & 96.8               & 98.5               \\
One-step  & 1               & DDIM       & 30               & 96.6               & 98.2               \\
Iterative & 400             & Iterative  & 400              & \textbf{98.3}               & \textbf{98.8}               \\
DDIM      & 30              & DDIM       & 30               & 98.2               & 98.6               \\ \bottomrule
\end{tabular}
\label{tab: training-inference consistency.}
\end{table}

\noindent\textbf{The effect of training-inference consistency.\quad}
In this paper, we train the reconstruction sub-network and segmentation sub-network in an end-to-end manner. During the training stage, the segmentation sub-network learns discrepancies between original inputs and their one-step denoised counterparts (rather than iterative or DDIM-denoised counterparts). Consequently, as shown in Table \ref{tab: training-inference consistency.}, using iterative or DDIM denoising during inference introduces distribution shifts of segmentation sub-network's inputs between the training and inference stages, which degrades anomaly detection accuracy and significantly slows down inference speed.

To maintain training-inference consistency for iterative or DDIM paradigms, the segmentation sub-network needs to learn from multi-step denoised results during training. As shown in Table \ref{tab: training-inference consistency.}, while this marginally improves performance, it slows down training by hundreds of times. Thus, our proposed one-step denoising framework achieves the optimal trade-off between speed and performance.

\begin{figure}[h]
  \centering
  \includegraphics[width=1.0\linewidth]{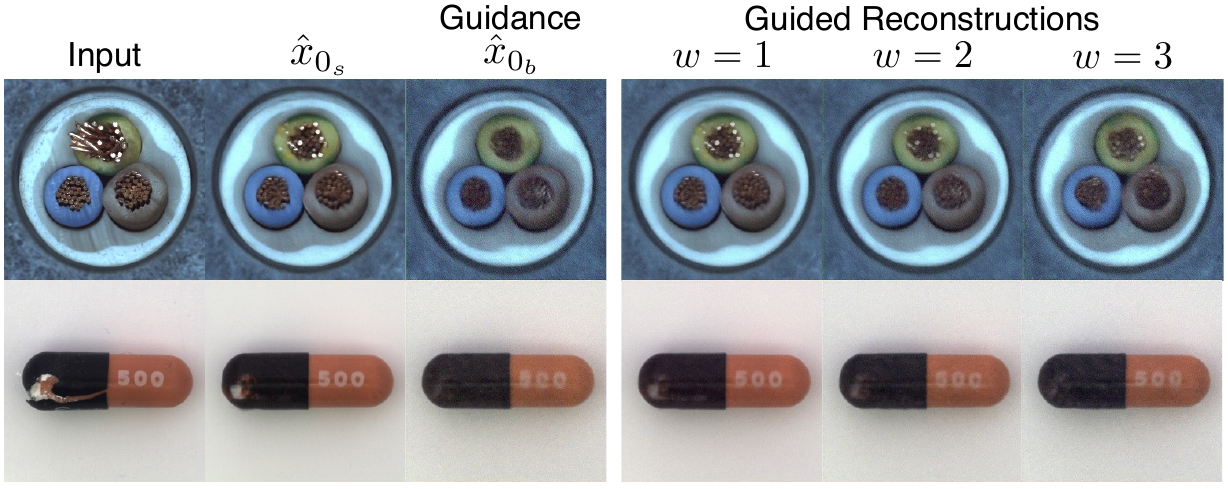}
  \caption{Ablations of guidance scale $w$.}
  \label{fig:guidance_scale_visual}
\end{figure}

\noindent\textbf{The effect of guidance scale.\quad}We further investigate how guidance information $\hat{x}_{0_b}$ modulates anomaly-free reconstruction through the guidance scale $w$. As shown in Figure \ref{fig:guidance_scale_visual}, the guided reconstructions increasingly resemble the guidance $\hat{x}_{0_b}$ as the guidance scale increases. In this paper, $w$ is empirically set to 1 to obtain the optimal fusion of reconstructions from two noise scales.

\section{Conclusion}
\label{sec:conclusion}
In this paper, we present DiffusionAD, a novel anomaly detection pipeline consisting of a reconstruction sub-network and a segmentation sub-network.
Firstly, we redefine the reconstruction process using a diffusion model, employing the noise-to-norm paradigm, where anomalous regions undergo perturbation by Gaussian noise and subsequently are reconstructed to appear normal.
Secondly, we revisit the entire denoising process and propose a one-step denoising paradigm, offering significant speed improvements over the iterative denoising approach while maintaining comparable reconstruction quality.
Furthermore, we introduce the norm-guided paradigm to leverage different noise scales' advantages in addressing the non-trivial challenge posed by various anomaly manifestations.
Finally, the segmentation sub-network leverages the inconsistencies and commonalities between the input image and its anomaly-free restoration to predict pixel-level anomaly scores.
Extensive evaluations across four datasets demonstrate that DiffusionAD surpasses current state-of-the-art methods, underscoring the efficacy and broad applicability of this novel pipeline.

\noindent\textbf{Limitations and Future Work.} While DiffusionAD demonstrates superior performance across multiple benchmarks, several limitations warrant further investigation. First,  the current norm-guided denoising paradigm relies on predefined noise scales, which may not optimally adapt to highly dynamic or context-dependent anomaly patterns.
Future work could explore adaptive noise scales and dynamic guidance based on the input image. 
Furthermore, the current anomaly synthesis strategy primarily relies on Perlin noise and texture replacements, which may not fully capture the diversity of real-world anomalies. 
Extending the anomaly synthesis to incorporate more complex and realistic anomaly patterns could further enhance the model's generalization capability.

\noindent\textbf{Broader Impact.} DiffusionAD shows potential applications across various industrial contexts.
Its real-time inference capability facilitates integration into existing production environments with less computational requirements. The approach handles diverse anomaly types, from texture irregularities to structural defects, making it suitable for industries with varying quality control requirements. These features position DiffusionAD as a practical solution for various industrial anomaly detection.

\bibliographystyle{IEEEtran}
\bibliography{main}

\begin{IEEEbiography}[{\includegraphics[width=1.2in,height=1.25in,clip,keepaspectratio]{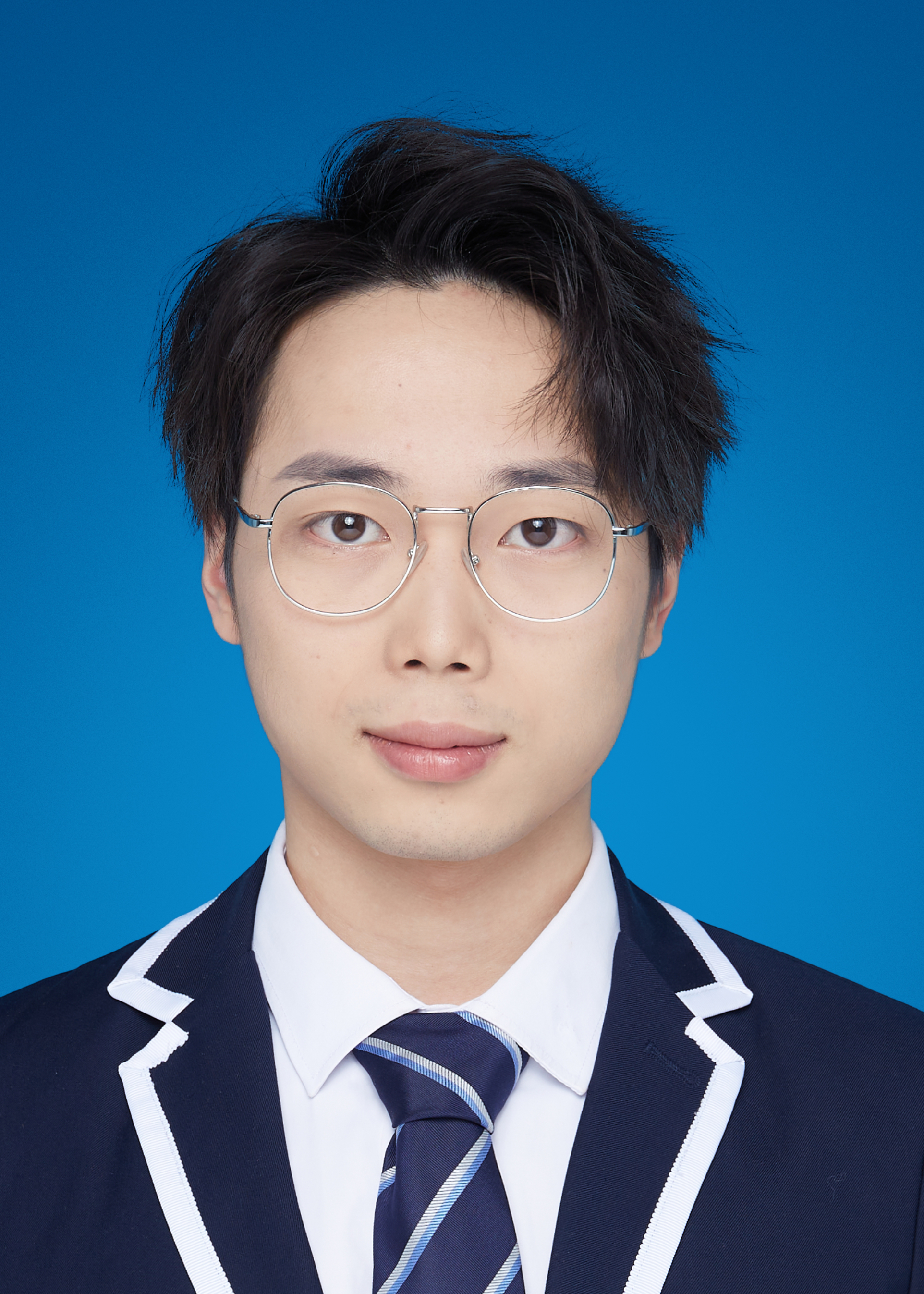}}]{Hui Zhang} is currently working toward the Ph.D. degree in Computer Science at Fudan University, advised by Prof. Zuxuan Wu and Prof. Yu-Gang Jiang. His research interests include generative models, anomaly detection, and large multimodal models.
\end{IEEEbiography}

\begin{IEEEbiography}[{\includegraphics[width=1.2in,height=1.25in, clip,keepaspectratio]{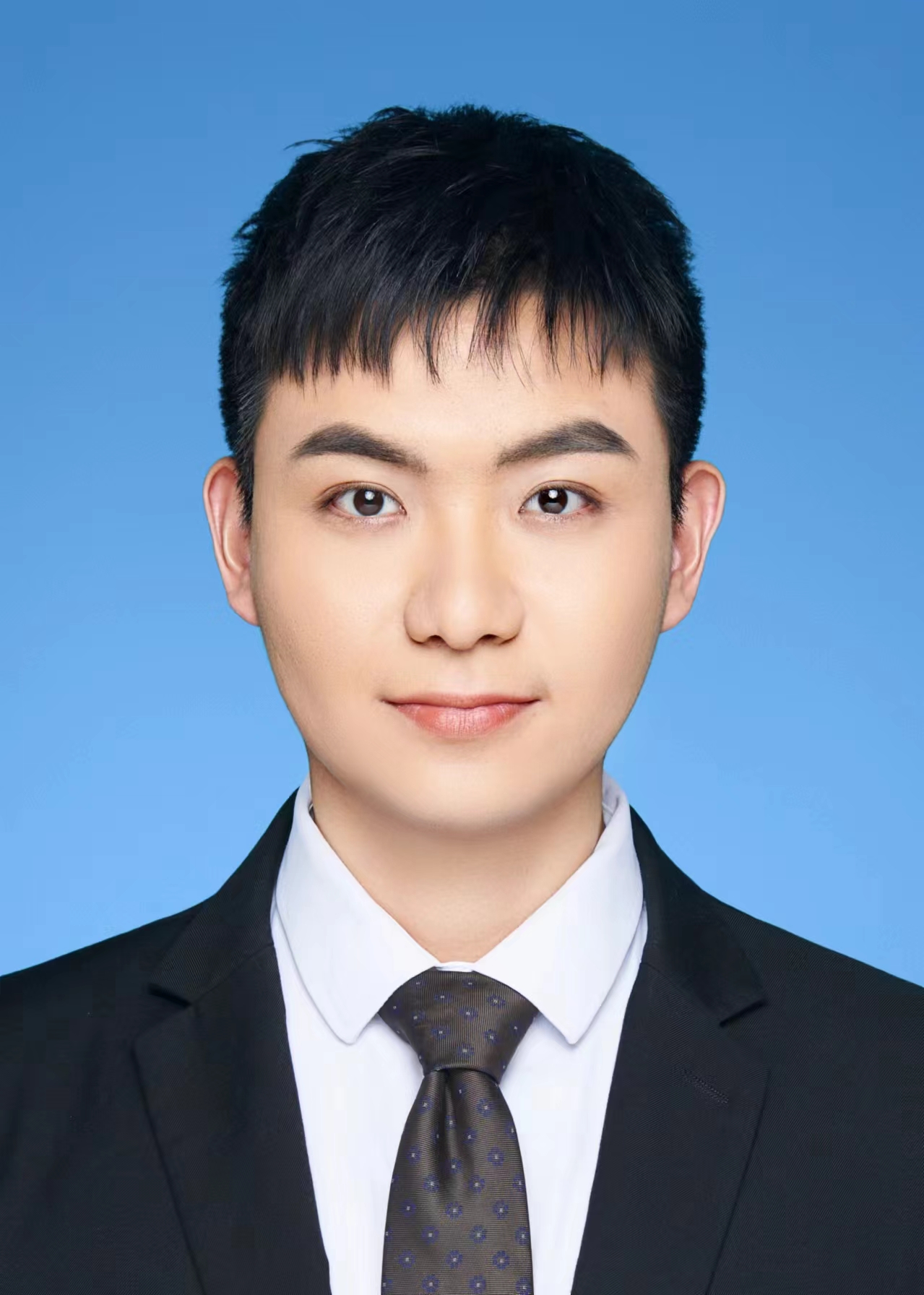}}]{Zheng Wang} received his Ph.D. in Computer Science from Fudan University with Prof. Yu-Gang Jiang in 2022. He is currently a lecturer in the School of Computer Science at Zhejiang University of Technology. His research interests include computer vision and multimedia learning. \end{IEEEbiography}

\begin{IEEEbiography}[{\includegraphics[width=1.2in,height=1.25in,clip,keepaspectratio]{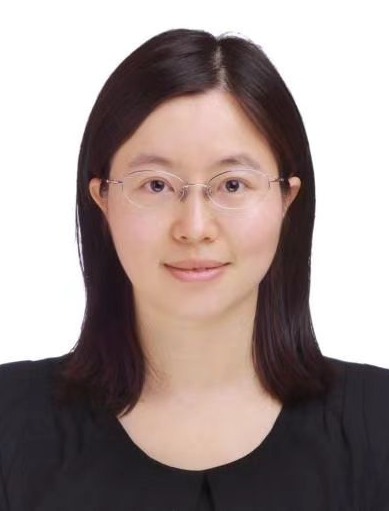}}]{Dan Zeng} (Senior Member, IEEE) received the Ph.D. degree in Circuits and Systems from the University of Science and Technology of China, Hefei. She is currently a Full Professor and the Dean of the Department of Communication Engineering, Shanghai University. Her main research interests include computer vision, multimedia analysis, and machine learning. She is a TC Member of IEEE MSA and IEEE MMSP. She is serving as a Vice Chair for IEEE COMSOC MMTC Exco Committee.
 \end{IEEEbiography}
 
\begin{IEEEbiography}[{\includegraphics[width=1.2in,height=1.25in,clip,keepaspectratio]{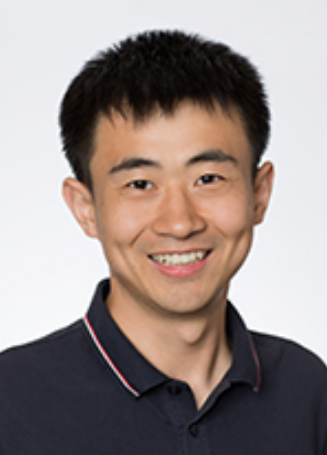}}]{Zuxuan Wu} 
received his Ph.D. in Computer Science from the University of Maryland with Prof. Larry Davis in 2020. He is currently an Associate Professor at the Institute of Trustworthy Embodied AI, Fudan University. His research interests are in computer vision and deep learning. His work has been recognized by an AI 2000 Most Influential Scholars Honorable Mention in 2021, a Microsoft Research PhD Fellowship (10 people Worldwide) in 2019 and a Snap PhD Fellowship (10 people Worldwide) in 2017. \end{IEEEbiography}

\begin{IEEEbiography}[{\includegraphics[width=1.2in,height=1.25in,clip,keepaspectratio]{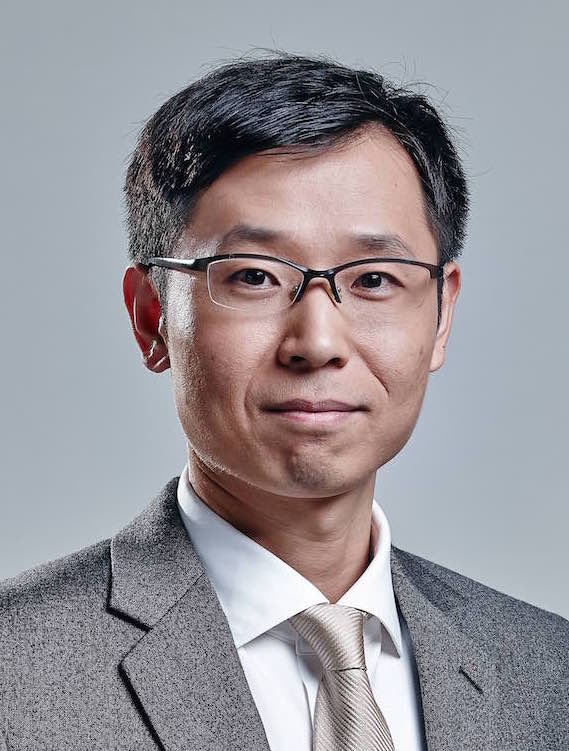}}]{Yu-Gang Jiang} (Fellow, IEEE) received the PhD degree in Computer Science from City University of Hong Kong in 2009 and worked as a Postdoctoral Research Scientist at Columbia University, New York, during 2009-2011. He is currently a Distinguished Professor at the Institute of Trustworthy Embodied AI, Fudan University, Shanghai, China. His research lies in the areas of multimedia, computer vision, embodied AI and trustworthy AI. His research has led to the development of innovative AI tools that have been used in many practical applications like defect detection for high-speed railway infrastructures. His open-source video analysis toolkits and datasets such as CU-VIREO374, CCV, THUMOS, FCVID and WildDeepfake have been widely used in both academia and industry. He currently serves as Chair of ACM Shanghai Chapter and Associate Editor of several international journals. For contributions to large-scale and trustworthy video analysis, he was elected to Fellow of IEEE, IAPR and CCF.
 \end{IEEEbiography}

\vfill

\end{document}